\documentclass[11pt,a4paper]{article}

\usepackage[T1]{fontenc}
\usepackage{lmodern}            % Latin Modern (clean, modern Computer Modern)
\usepackage[scaled=0.85]{beramono} % Nicer monospace
\usepackage{microtype}          % Optical margin alignment & font expansion
\usepackage{longtable}
\usepackage[
  hmargin=3.8cm,
  vmargin=3cm
]{geometry}

\usepackage{parskip}            % Paragraph spacing instead of indent
\usepackage{enumitem}           % Fine-grained list control
\setlist{
  topsep=0.5em,
  itemsep=0.2em,
  parsep=0pt
}

\usepackage{fancyhdr}
\fancypagestyle{plain}{%         % First page style
  \fancyhf{}
  
  \fancyfoot[C]{\small\thepage}
}

\usepackage{titlesec}

\titleformat{\section}
  {\large\sffamily\bfseries}
  {\thesection}{0.75em}{}

\titleformat{\subsection}
  {\normalsize\sffamily\bfseries}
  {\thesubsection}{0.75em}{}

\titlespacing*{\section}{0pt}{2.5ex}{1.2ex}
\titlespacing*{\subsection}{0pt}{2ex}{1ex}

\usepackage{xcolor}
\definecolor{accent}{HTML}{2563EB}    % A clean blue
\definecolor{textcodered}{rgb}{0.55,0.1,0.1}

\usepackage[
  colorlinks = true,
  urlcolor = accent,
  citecolor = accent,
  linkcolor = accent
]{hyperref}

\usepackage{graphicx}
\usepackage{booktabs}
\usepackage{caption}
\usepackage{tikz}
\usetikzlibrary{arrows.meta, positioning, shapes.geometric, fit, backgrounds}
\usepackage{float}  % For [H] placement
\usepackage{wrapfig}

\usepackage[round,authoryear]{natbib}

\usepackage{mdframed}
\renewenvironment{abstract}{%
  \begin{mdframed}[
    linewidth=0pt,
    backgroundcolor=black!4,
    innertopmargin=12pt,
    innerbottommargin=12pt,
    innerleftmargin=14pt,
    innerrightmargin=14pt,
    roundcorner=3pt,
    skipabove=\baselineskip,
    skipbelow=\baselineskip
  ]
  \noindent\small
}{%
  \end{mdframed}
}

\usepackage{cite}
\usepackage{amsmath,amssymb,amsfonts}
\usepackage{algorithmic}
\usepackage{graphicx}
\usepackage{textcomp}
\usepackage{xcolor}
\usepackage{booktabs}
\usepackage{multirow}
\usepackage{array}
\usepackage{hyperref}
\usepackage{tikz}
\usepackage{pgfplots}
\usepackage{subcaption}
\usepackage{caption}
\pgfplotsset{compat=1.17}
\usetikzlibrary{shapes.geometric, arrows.meta, positioning, fit, backgrounds, calc, decorations.pathreplacing}

\begin{document}

\title{%
  \vspace{-1.5cm}%
  {\Huge\texttt{\textbf{SR-Platform}}}\\[6pt]
  {\large\textsf{An Agentic Pipeline for Natural Language-Driven Robot Simulation Environment Synthesis}}\\[3pt]
  {\normalsize\href{https://strikerobot.ai/}{\textsf{strikerobot.ai}}}%
  \vspace{-0.5cm}%
}
\author{%
  \normalsize Ben Wei Lim\textsuperscript{1} \quad
  Minh Duc Le\textsuperscript{1} \quad Thang Truong\textsuperscript{1} \quad
  Thanh Nguyen Canh\textsuperscript{1}\\[4pt]
  \small\textsuperscript{1}Strike Robotics 
  % \quad   \textsuperscript{2}Sorbonne University 
  \\
  \texttt{contact@strikerobot.ai}
}
\date{}

\maketitle
\begin{figure}[!ht]
    \centering
    \includegraphics[width=\linewidth]{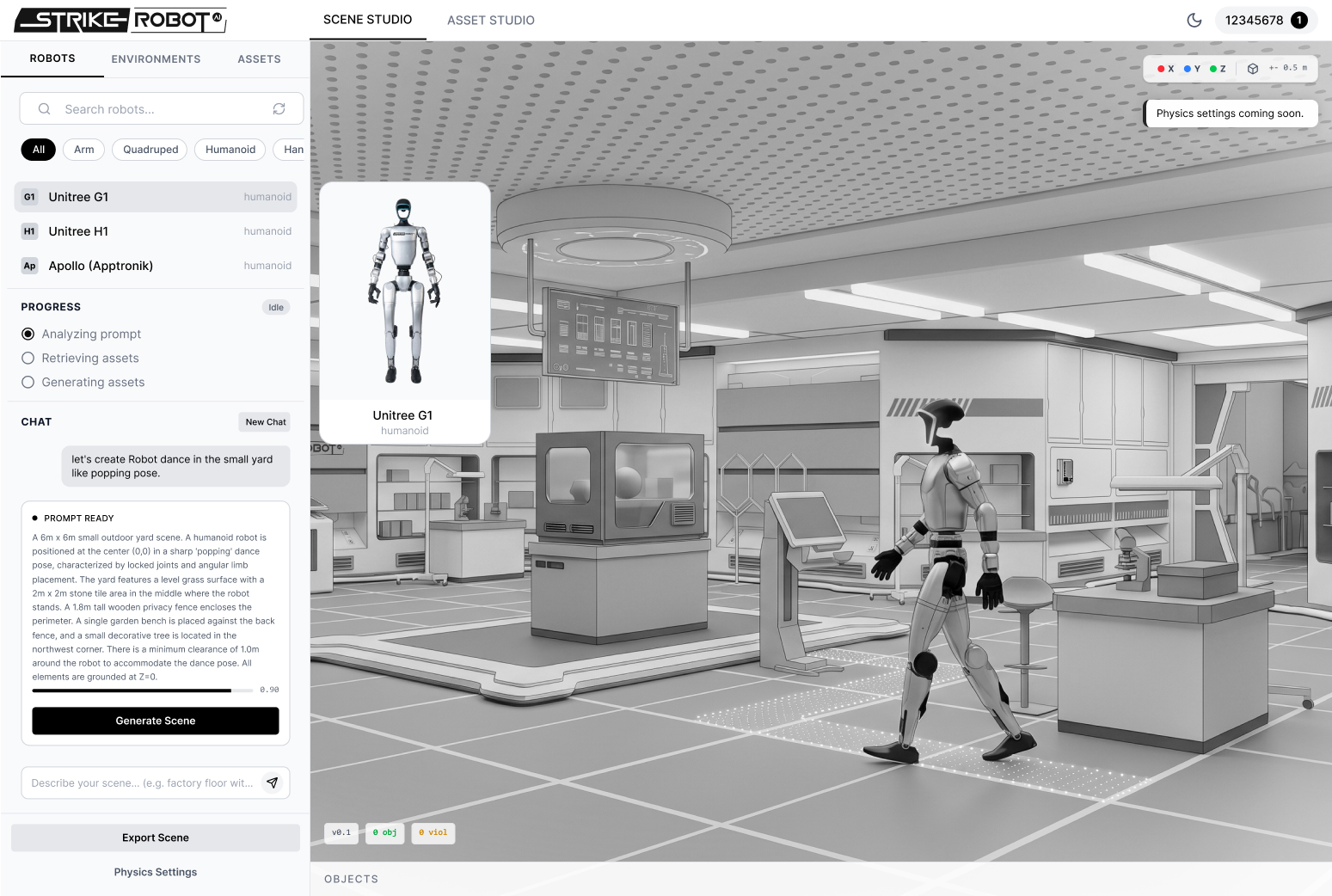}
    \caption{SR-Platform web interface generating a robot simulation environment from a natural-language scene prompt. The interface combines prompt refinement, robot selection, real-time generation progress, and browser-based MuJoCo scene visualization.}
    \label{fig:interface}
\end{figure}

\begin{abstract}
Generating robot simulation environments remains a major bottleneck in simulation-based robot learning. Constructing a training-ready MuJoCo scene typically requires expertise in 3D asset modeling, MJCF specification, spatial layout, collision avoidance, and robot-model integration. We present SR-Platform, a production-deployed agentic system that converts free-form natural language descriptions into executable, physically valid MuJoCo environments. SR-Platform decomposes scene synthesis into four stages: an LLM-based orchestrator that converts user intent into a structured scene plan; an asset forge that retrieves cached assets or generates new 3D geometry through LLM-to-CadQuery synthesis; a layout architect that assigns object poses and verifies industrial constraints; and a bridge layer that assembles the final MJCF scene and merges the selected robot model. The system is deployed as a nine-service Docker stack with WebSocket progress streaming, MinIO-backed mesh storage, Qdrant-based semantic asset retrieval, Redis job state, and InfluxDB telemetry. Using 30 days of production telemetry covering 611 successful LLM calls, SR-Platform generates five-object scenes with a median end-to-end latency of approximately 50 s, while cache-accelerated scenes complete in approximately 30--40 s. The asset forge shows an 11.3\% first-attempt retry rate with automatic recovery, and cached asset retrieval removes per-object LLM calls for previously generated object types. These results show that agentic scene synthesis can reduce the manual effort required to create diverse robot training environments, enabling users to produce executable MuJoCo scenes from plain English prompts in under one minute.
\end{abstract}

% \begin{keywords}
% robot simulation, scene synthesis, natural language interfaces, agentic AI, MuJoCo, procedural generation, sim-to-real
% \end{keywords}

%==============================================================================
\section{Introduction}
\label{sec:introduction}
%==============================================================================

Robot learning at scale requires large collections of diverse, physically plausible simulation environments. Reinforcement learning, imitation learning, manipulation benchmarking, and sim-to-real evaluation all depend on environments that reflect the geometry, object diversity, and spatial constraints of the target deployment domain~\citep{matas2018sim,zhao2020sim}. In practice, however, constructing these environments remains a major bottleneck. A robotics engineer must locate or author 3D assets, convert them into simulator-compatible mesh formats, write a valid MJCF scene description, position objects without collisions, satisfy workspace and clearance constraints, and merge a robot model with an appropriate spawn pose. This process requires expertise in robotics, 3D modeling, and simulator-specific file formats, making it difficult to scale environment generation to the diversity required for robust policy training.

MuJoCo has become a widely used simulator for robot learning because it provides efficient rigid-body dynamics, articulated-body simulation, and a compact XML-based modeling format~\citep{todorov2012mujoco}. However, the MJCF representation is designed for precision and simulator control, not for natural authoring by non-specialist users. A user may be able to describe a desired workspace in plain English, such as an assembly cell, warehouse aisle, laboratory bench, or kitchen-like manipulation environment, but this description must still be transformed into structured objects, geometric assets, valid spatial layouts, and simulator-executable XML. The gap between a natural-language scene description and a runnable robot simulation therefore, remains substantial.

Recent advances in large language models (LLMs) suggest a promising path toward automating parts of this workflow. LLMs have demonstrated strong capabilities in structured output generation~\citep{brown2020language}, code synthesis~\citep{chen2021evaluating}, and embodied or spatial reasoning~\citep{driess2023palm}. In robotics, LLMs have been used to generate reward functions~\citep{ma2023eureka}, infer task plans~\citep{ahn2022saycan}, produce robot-control programs~\citep{liang2023code}, and synthesize task-environment descriptions~\citep{wang2023robogen}. These systems show that language models can provide useful semantic reasoning for robotics workflows. However, most prior approaches either assume a fixed object library, stop at symbolic scene or task descriptions, or require manual intervention before the result can be executed in a physics simulator. As a result, the end-to-end problem of converting an unconstrained language prompt into a physically valid, simulator-ready robot scene remains insufficiently addressed.

To address these challenges, this paper presents SR-Platform, a production-deployed agentic system for natural-language-driven robot simulation environment synthesis. Given a free-form workspace description, SR-Platform generates an executable MuJoCo scene with retrieved or synthesized assets, spatially assigned object poses, an integrated robot model, and spawn coordinates. Rather than using an LLM as a monolithic scene generator, SR-Platform embeds LLM calls within a typed, auditable, and cache-aware pipeline that separates semantic planning, asset generation, layout reasoning, and simulator assembly. Requested objects are first matched against a vector database of previously generated assets; if no suitable match is found, an LLM synthesizes CadQuery code to generate a new simulator-compatible mesh, which is then indexed for future reuse. Industrial validity is also checked during layout generation through constraints such as electrical clearance, fire-safety egress, machinery safety distance, and ventilation clearance. As shown in Fig.~\ref{fig:interface}, a user describes an industrial workspace through a web interface and receives a generated scene that can be visualized directly in the browser. The system is evaluated using 30 days of production telemetry covering 611 successful LLM calls, with additional object-generation benchmarks and model-routing results reported in the appendix. The main contributions of this paper can be organized as follows:

\begin{enumerate}
    \item An end-to-end agentic pipeline is introduced for converting natural-language workspace descriptions into executable MuJoCo simulation scenes with robot models, mesh assets, object poses, and spawn coordinates.
    \item A cache-aware asset forge is presented that combines semantic retrieval with LLM-to-CadQuery generation, reducing redundant LLM calls for previously generated object types.
    \item Industrial layout verification is integrated into the scene synthesis process, enabling generated layouts to be checked against clearance and safety constraints before MJCF assembly.
    \item Production telemetry from 611 successful LLM calls is reported, including per-stage latency, retry behavior, throughput limits, and cache-accelerated end-to-end generation estimates.
\end{enumerate}

%==============================================================================
\section{Related Work}
\label{sec:related}
%==============================================================================

\subsection{Simulation Environment Generation}

Simulation environment generation has long been studied as a way to reduce the cost of robot learning and evaluation. Existing platforms provide either high-fidelity simulation backends or curated environment collections, but they often require substantial manual effort to construct new scenes. ThreeDWorld provides an interactive physical simulation environment with rich multimodal perception, but scene authoring remains largely manual~\citep{gan2020threedworld}. BEHAVIOR-1K introduces a large set of household activities and corresponding environments for embodied AI, but its scope is tied to a predefined benchmark and simulator-specific assets~\citep{li2023behavior}. RoboGen uses generative models to create task-environment pairs for robot learning, but still relies on curated object resources and does not focus on production deployment for industrial scene synthesis~\citep{wang2023robogen}. In contrast, SR-Platform targets on-demand generation of industrial robot workspaces from natural language and produces directly executable MuJoCo scenes rather than only task descriptions or benchmark environments.

\subsection{Language Models for Robotics}
Large language models have increasingly been used as semantic reasoning components in robotics systems. SayCan grounds language instructions in robot affordances to select feasible actions~\citep{ahn2022saycan}, while Code as Policies uses language models to generate robot-control programs from natural-language instructions~\citep{liang2023code}. Eureka demonstrates that LLMs can synthesize reward functions for reinforcement learning~\citep{ma2023eureka}, and Voyager shows how an LLM-based agent can perform open-ended exploration through iterative skill acquisition~\citep{wang2023voyager}. These systems show that LLMs can support planning, control, reward design, and embodied decision-making. SR-Platform addresses a different but complementary problem: instead of using language models to control a robot inside an existing environment, language models are used to construct the environment itself, including the scene plan, assets, layout, and simulator-ready MJCF representation.

\subsection{LLM-Based Code and Asset Generation}
Code generation models have shown strong performance in producing structured programs from natural-language specifications~\citep{chen2021evaluating}. This capability is useful for procedural asset generation because CAD tools expose programmatic interfaces that can be driven by generated code. CadQuery provides a Python-based parametric CAD interface, making it suitable for LLM-generated geometry when the target output must be physically meaningful rather than only visually plausible~\citep{cadquery2023}. Text-to-3D methods such as DreamFusion generate 3D objects from language prompts, but these approaches primarily optimize visual appearance and are not always suited for watertight, scale-consistent, simulator-compatible geometry~\citep{poole2022dreamfusion}. SR-Platform instead uses LLM-to-CadQuery synthesis to generate STL assets that can be inserted into physics simulation, combined with validation, retry, and fallback mechanisms to improve robustness.

\subsection{Agentic and Tool-Using AI Systems}
Agentic AI systems decompose complex tasks into tool-mediated reasoning steps. ReAct introduced a framework in which language models interleave reasoning and action, enabling them to call external tools during problem solving~\citep{yao2023react}. LangGraph extends this idea with graph-based orchestration for stateful multi-step LLM applications~\citep{langchain2024langgraph}. SR-Platform follows this agentic design principle, but applies it to robot simulation synthesis rather than conversational question answering or robot action selection. The scene-generation process is divided into explicit stages for orchestration, asset resolution, layout design, and MJCF assembly. This modular structure makes the system easier to cache, debug, validate, and deploy than a single end-to-end LLM prompt.

\subsection{Sim-to-Real Transfer and Environment Diversity}
The sim-to-real gap remains a central challenge in robot learning~\citep{zhao2020sim}. Domain randomization has been widely used to improve transfer by varying visual appearance, object placement, lighting, and dynamics during simulation~\citep{tobin2017domain}. However, visual randomization alone does not solve the problem of structural diversity: robots also need to experience different workspace layouts, object combinations, and task-relevant spatial constraints. SR-Platform focuses on this structural aspect of diversity by generating new environment configurations from natural language. High-fidelity simulators and robotics frameworks such as MuJoCo, Newton, and Isaac Lab provide important physics backends for robot learning~\citep{todorov2012mujoco,mittal2023orbit,nvidia2026newton}, while SR-Platform operates at the environment-synthesis layer above the simulator, producing executable scenes that can be consumed by such backends.
%==============================================================================
\section{System Architecture}
\label{sec:architecture}
%==============================================================================

\subsection{Overview}

\begin{figure}[t]
\centering
\includegraphics[width=0.85\textwidth]{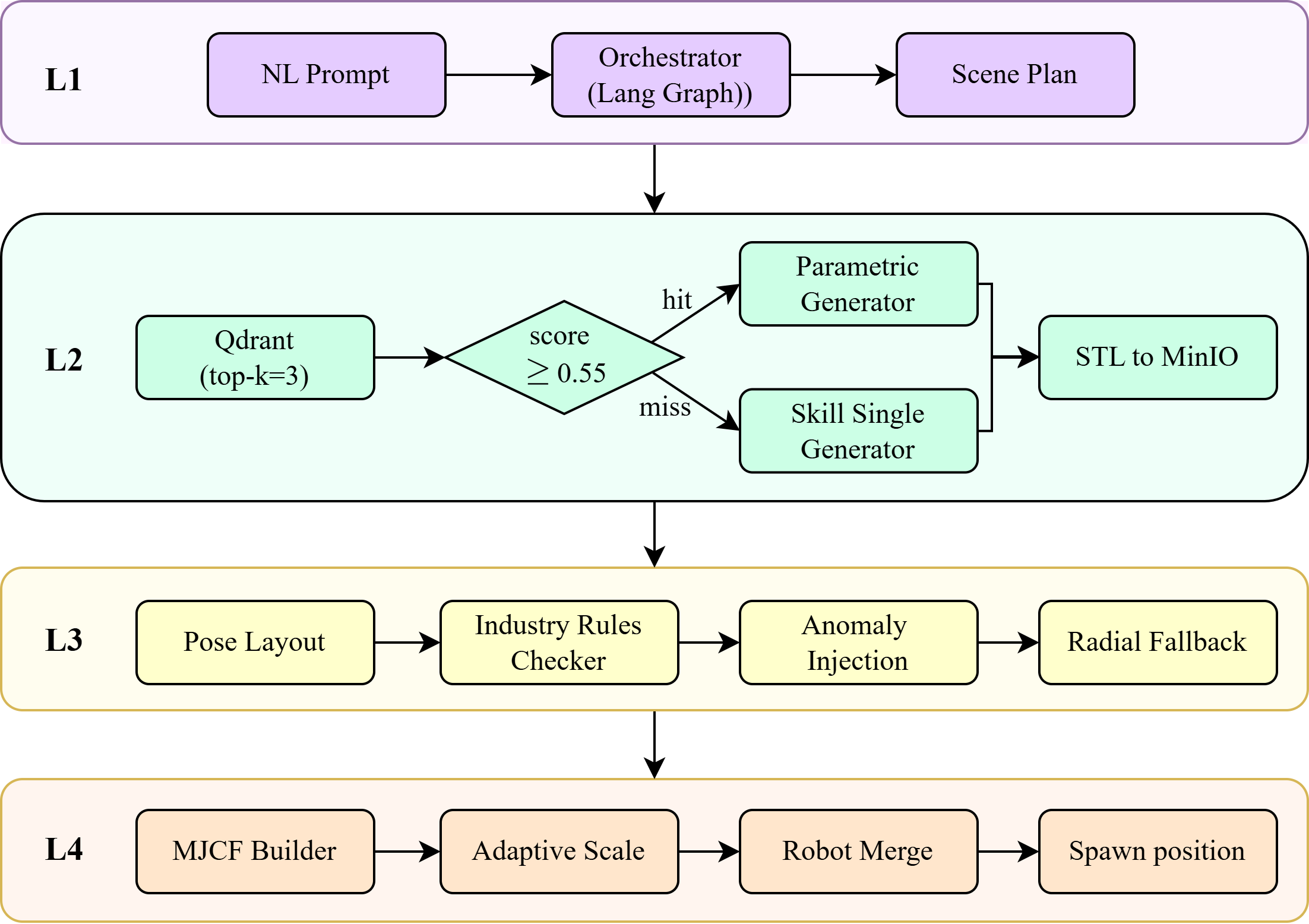}
\caption{Four-layer SR-Platform pipeline for natural-language-driven robot simulation synthesis. L1 parses the user prompt into a structured scene plan; L2 retrieves cached assets from Qdrant or generates new STL meshes through LLM-to-CadQuery synthesis; L3 computes spatial layout and verifies industrial constraints; and L4 assembles the final MJCF scene and merges the selected robot model.}
\label{fig:pipeline}
\end{figure}

SR-Platform is organized as a four-layer scene-synthesis pipeline coordinated by a \texttt{PipelineCoordinator}. Each stage receives a structured \texttt{PipelineContext}, enriches it with stage-specific outputs, and passes the updated context to the next stage. This design keeps semantic planning, asset resolution, spatial layout, and MJCF assembly separate, making the system easier to validate, cache, debug, and extend.
Fig.~\ref{fig:pipeline} shows the complete L1--L4 architecture. The L1 orchestrator converts a natural-language prompt into a structured scene plan. The L2 asset forge resolves each requested object into a simulator-compatible mesh, either by retrieving a cached asset or by generating new geometry. The L3 layout architect computes object poses and checks industrial constraints. Finally, the L4 bridge assembles the MJCF scene and merges the selected robot model.
The system is deployed as a production Docker stack composed of backend, worker, frontend, Qdrant, PostgreSQL, Redis, MinIO, InfluxDB, and pgAdmin services. The frontend communicates with the backend through a WebSocket interface, allowing progress updates to be streamed while long-running generation jobs execute asynchronously.

\subsection{L1: Orchestrator}

The L1 orchestrator is responsible for converting a free-form user prompt into a structured scene plan. The input is a natural-language description of a robot workspace, such as an assembly line, warehouse aisle, laboratory bench, or kitchen-like manipulation scene. The output is a JSON plan containing the room dimensions, object list, semantic object labels, selected robot identity, and optional anomaly-generation parameters.
The orchestrator is implemented as a lightweight LangGraph workflow with two main steps: an LLM call for scene-plan generation and a parsing step that validates the returned JSON structure. This stage does not create geometry or assign final object poses. Instead, it produces a compact symbolic representation that downstream stages can consume. Separating this planning step from asset generation and layout reasoning reduces the complexity of each LLM call and makes failures easier to localize.
The L1 output acts as the primary data contract for the rest of the pipeline. Each object entry contains enough semantic information for the asset forge to retrieve or generate a corresponding mesh, while the room and robot fields provide the layout architect and bridge layer with global scene constraints.

\begin{figure}
    \centering
    \includegraphics[width=\linewidth]{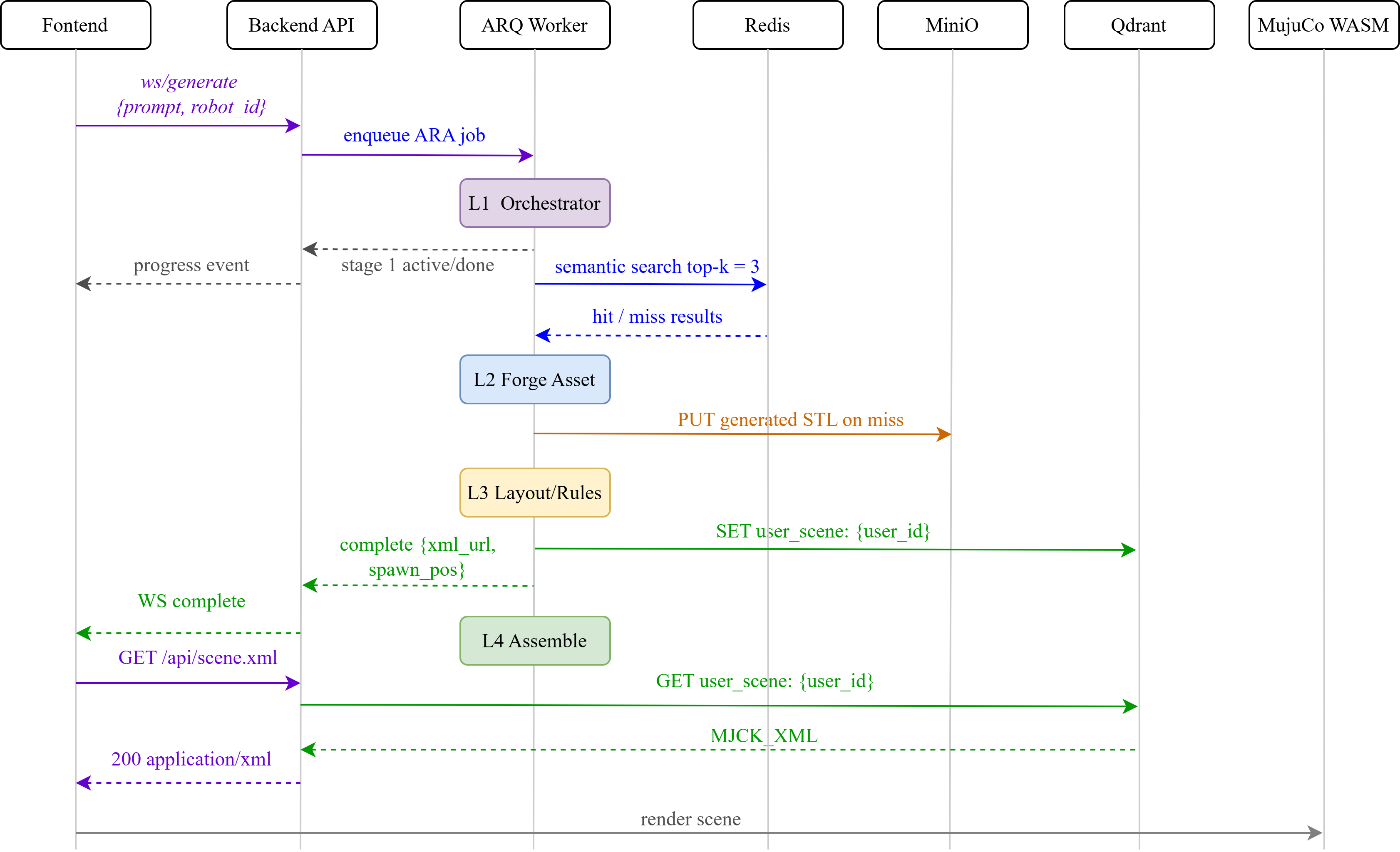}
    \caption{End-to-end request lifecycle for scene generation. A WebSocket request is submitted from the frontend, queued through ARQ, processed by the L1--L4 pipeline, connected to Qdrant for semantic asset retrieval, MinIO for generated mesh storage, and Redis for per-user scene state, then returned as MJCF XML for browser rendering through MuJoCo WASM.}
    \label{fig:request_lifecycle}
\end{figure}

\subsection{L2: Asset Forge}

The L2 asset forge resolves each object in the scene plan into a 3D mesh that can be inserted into the simulator. For every requested object label, the forge first performs semantic retrieval against a Qdrant vector database. Object labels are embedded using a high-dimensional text embedding model, and the top candidate assets are retrieved from the asset collection. If the best match exceeds the similarity threshold, the corresponding STL mesh is loaded from MinIO and reused.
If no suitable cached asset is found, the forge invokes an LLM-based generation agent to synthesize CadQuery code for the requested object. The generated code is executed to produce an STL mesh, which is then stored in MinIO and indexed into Qdrant for future reuse. This cache-first strategy is important because asset generation is the most expensive stage of the pipeline: every cache miss requires a separate LLM-to-CAD generation call, while cache hits bypass this call entirely.
The asset forge may also apply validation and retry mechanisms. Generated meshes can be checked for geometric validity before being accepted, and failed CadQuery programs can be regenerated automatically. This design allows the system to recover from malformed code, invalid geometry, or overly complex object descriptions without failing the entire scene-generation request.

\subsection{L3: Layout Architect}

The L3 layout architect assigns spatial poses to all resolved objects. Its input is the enriched scene plan from L2, including object identities, mesh metadata, and room dimensions. Its output is a laid-out scene plan containing positions, rotations, and constraint-checking results.
This stage uses an LLM-based layout model to propose object placements that are semantically plausible and physically consistent. For example, tables should stand on the floor, control panels should be placed near walls or machinery, and objects should not be placed outside the room boundary. The layout architect also checks for basic physical validity, such as object overlap and clearance from walls.
Industrial validity is incorporated through an \texttt{IndustryRulesChecker}. The checker evaluates generated layouts against constraints inspired by electrical clearance, fire-safety egress, machinery safety distance, and ventilation clearance requirements. Detected violations are returned as structured entries, allowing the system to resample layouts, notify the user, or generate anomaly-labeled scenes when anomaly injection is requested.

\subsection{L4: Bridge and MJCF Assembly}

The L4 bridge converts the laid-out scene plan into an executable MuJoCo scene. This stage constructs the floor, walls, object bodies, mesh references, materials, and global simulation settings in MJCF format. Each generated or retrieved STL asset is inserted as a mesh-backed body with its position and rotation determined by the L3 layout output.
The bridge also performs robot integration. A \texttt{RobotRegistry} resolves the selected robot model, and a \texttt{RobotMerger} inserts the robot MJCF into the generated environment at the computed spawn position. Coordinate-frame consistency is handled at this stage so that the generated scene can be rendered correctly in the browser and executed in MuJoCo. Because L4 relies on local XML construction and mesh metadata rather than LLM calls, it is designed to be deterministic and fast relative to the planning and generation stages.
The final output of L4 includes the complete MJCF XML, a scene-only MJCF variant when needed, and robot spawn coordinates. These outputs are stored as per-user scene state and returned to the frontend for visualization.

\subsection{Request Lifecycle and Deployment}

Fig.~\ref{fig:request_lifecycle} shows how a scene-generation request moves through the deployed system. A user submits a prompt through the frontend WebSocket interface. The backend authenticates the request, enqueues a generation job through ARQ, and streams progress events back to the frontend as each pipeline stage becomes active or completes.
During execution, the worker runs the L1--L4 pipeline. Qdrant is queried during asset retrieval, MinIO stores and serves generated STL meshes, Redis stores job state and per-user scene state, and PostgreSQL maintains persistent metadata such as users, prompts, assets, and API-key configuration. InfluxDB records telemetry for LLM calls and system monitoring. After L4 assembly completes, the frontend retrieves the MJCF scene and renders it through MuJoCo WASM and Three.js.
This architecture separates interactive user communication from long-running generation work. WebSocket streaming provides real-time feedback, while the asynchronous job queue prevents a single request from blocking the backend. Per-user job limits and worker concurrency settings are used to preserve fairness under load, and additional workers can be launched to increase throughput without changing the pipeline interface.

%==============================================================================
\section{Experimental Evaluation}
\label{sec:evaluation}
%==============================================================================

\subsection{Experimental Setup}
SR-Platform was evaluated using production telemetry collected over a 30-day deployment window. The dataset contains 611 successful LLM calls recorded from the \texttt{llm\_call} measurement in the InfluxDB \texttt{aes\_metrics} bucket. During this measurement period, the generation stages were routed through \texttt{venice/deepseek-v4-flash}, while asset-retrieval embeddings were generated with \texttt{venice/text-embedding-3-large}. The CadQuery generation timeout was set to \texttt{CQ\_LLM\_TIMEOUT=120s}.

The evaluation focuses on four operational questions: how much latency is introduced by each LLM-dependent stage, how end-to-end latency changes with object count and cache reuse, how reliably the asset generator recovers from failures, and how many concurrent scene-generation jobs the deployed system can support. Because direct wall-clock instrumentation for each pipeline stage was not yet fully connected during the measurement window, end-to-end latency is estimated from observed LLM-call distributions and known pipeline structure. This limitation is discussed explicitly in Section~\ref{sec:discussion}.

\subsection{Per-Stage Latency}

Table~\ref{tab:llm_latency} summarizes LLM call latency by pipeline stage. The L2 asset forge is the dominant latency contributor because each cache-miss object requires a separate LLM-to-CadQuery generation call. The \texttt{skill\_generator} calls show a median latency of 17.9~s and a p95 latency of 65.4~s. Retry calls are slower, with a median latency of 24.4~s and a p95 latency of 77.9~s, reflecting the additional difficulty of correcting malformed or invalid geometry.

The \texttt{unknown} tag contains L1 orchestration and L3 layout calls that were not separately labeled due to an instrumentation gap. These calls have a median latency of 15.9~s and a p95 latency of 73.9~s. One extreme outlier of 761.4~s was observed, but the p95 value is more representative for system design. L4 bridge assembly is not included in the LLM-call table because it performs local XML construction and mesh processing without invoking an LLM.

\begin{table}[t]
\centering
\caption{LLM call latency by pipeline stage measured from 30 days of production telemetry. The asset forge dominates runtime because each cache-miss object requires an LLM-generated CadQuery program, while L4 assembly is local XML and mesh processing with no LLM call.}
\label{tab:llm_latency}
\small
\begin{tabular}{lccccc}
\toprule
\textbf{Stage (tag)} & \textbf{Role} & \textbf{N} & \textbf{p50 (s)} & \textbf{p95 (s)}&\textbf{Max (s)} \\
\midrule
\texttt{skill\_generator}        & L2 CadQuery gen     & 300 & 17.9 & 65.4 & 114.2\\
\texttt{skill\_generator\_retry} & L2 retry            &  34 & 24.4 & 77.9 & 98.4\\
\texttt{validator\_retry} & L2 retry            &  1 & 12.7 & 12.7 & 12.7\\
\texttt{unknown}                 & L1 Orch + L3 Arch   & 275 & 15.9 & 73.9 & 761.4\\
\bottomrule
\end{tabular}
\end{table}

\subsection{End-to-End Scene Generation Latency}

Table~\ref{tab:e2e_latency} reports estimated end-to-end latency for representative scene-generation scenarios. For a scene with $N$ cache-miss objects, the pipeline requires one L1 call, $N$ L2 asset-generation calls, one L3 layout call, and no LLM call in L4. Since L2 processes objects in parallel up to the configured worker limit, the approximate end-to-end latency can be written as:
\[
T_{\mathrm{E2E}} \approx T_{L1} + 
\left\lceil \frac{N_{\mathrm{miss}}}{W} \right\rceil T_{L2}
+ T_{L3} + T_{L4},
\]
where $N_{\mathrm{miss}}$ is the number of cache-miss objects and $W$ is the maximum number of concurrent CadQuery-generation workers.

For a five-object scene in which all objects miss the cache, the estimated median end-to-end latency is approximately 50~s. For an eight-object scene, latency increases to approximately 68~s because the asset forge requires two generation batches when the worker limit is five. In contrast, when all requested objects are retrieved from the semantic cache, L2 generation calls are skipped and the median latency falls to approximately 30--40~s.

\begin{table}[t]
\centering
\caption{Estimated end-to-end scene generation latency under different cache and object-count scenarios. Cache misses require one L1 call, one L3 call, and one L2 asset-generation call per missing object, while full cache hits skip L2 generation and substantially reduce median latency.}
\label{tab:e2e_latency}
\small
\begin{tabular}{lccc}
\toprule
\textbf{Scenario} & \textbf{LLM Calls} & \textbf{E2E p50 (s)} & \textbf{E2E p95 (s)} \\
\midrule
L1 Orchestrator        & 1        &  $\sim$16 & $\sim$74  \\
L2 Forge               & 5        &  $\sim$18 & $\sim$65  \\
L2 Forge               & 8        &  $\sim$36 & $\sim$130  \\
L3 Architect           & 1        &  $\sim$16 & $\sim$74  \\
L4 Bridge              & 0        &  $<$1 & $<$2  \\
\midrule
5 objects, cache miss  & 7  (1+5+1) & $\sim$50  & $\sim$215 \\
8 objects, cache miss  & 10 (1+8+1) & $\sim$68  & $\sim$280 \\
All objects, cache hit & 2  (1+0+1) & $\sim$32  & $\sim$148 \\
\bottomrule
\end{tabular}
\end{table}

\begin{figure}[!ht]
    \centering
    \begin{subfigure}{0.49\linewidth}
        \includegraphics[width=\linewidth]{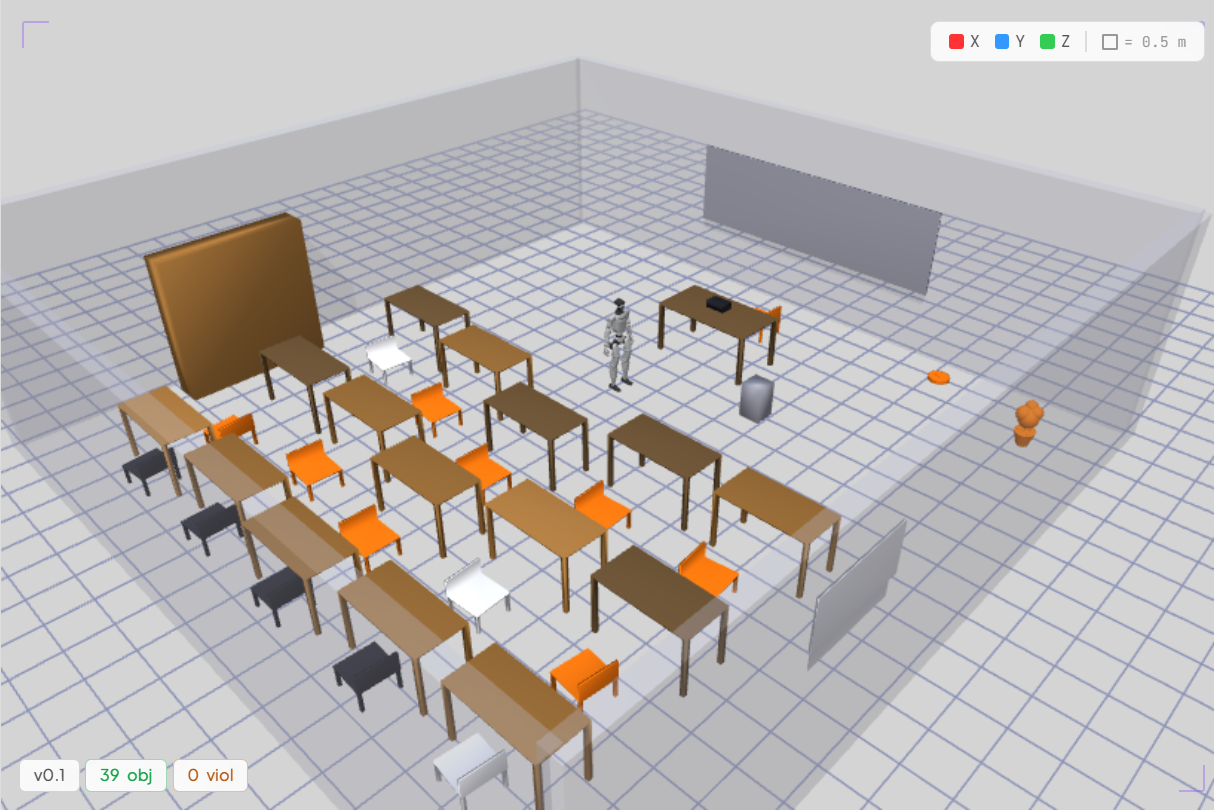}
        \caption{Classroom}
    \end{subfigure}
    \begin{subfigure}{0.49\linewidth}
        \includegraphics[width=\linewidth]{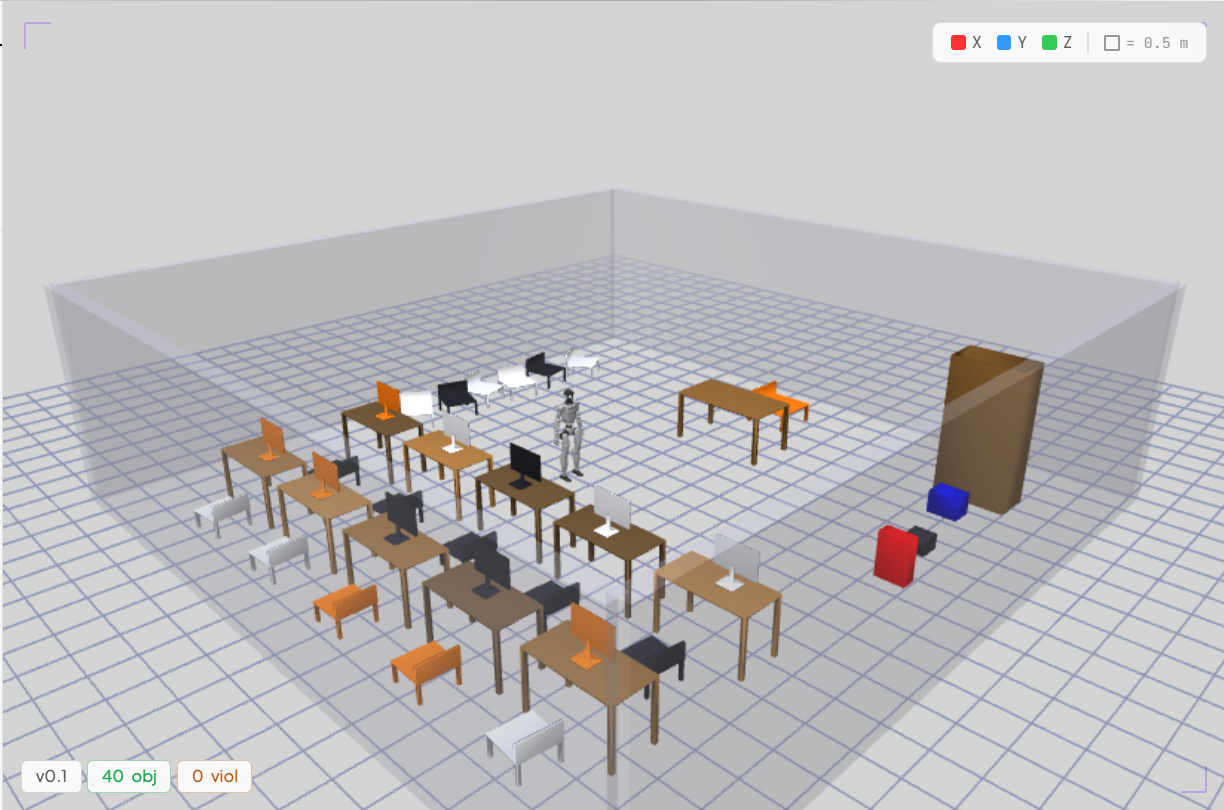}
        \caption{Computer class}
    \end{subfigure}
    \begin{subfigure}{0.49\linewidth}
        \includegraphics[width=\linewidth]{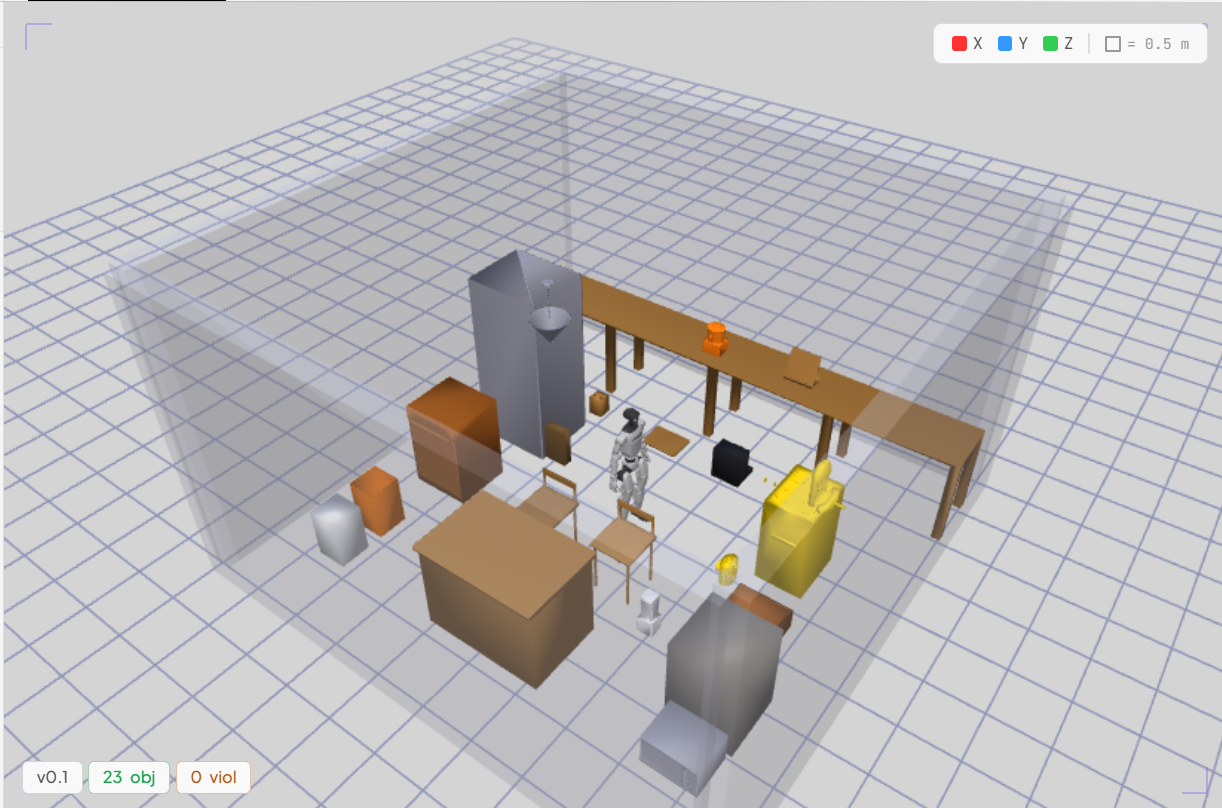}
        \caption{Kitchen}
    \end{subfigure}
    \begin{subfigure}{0.49\linewidth}
        \includegraphics[width=\linewidth]{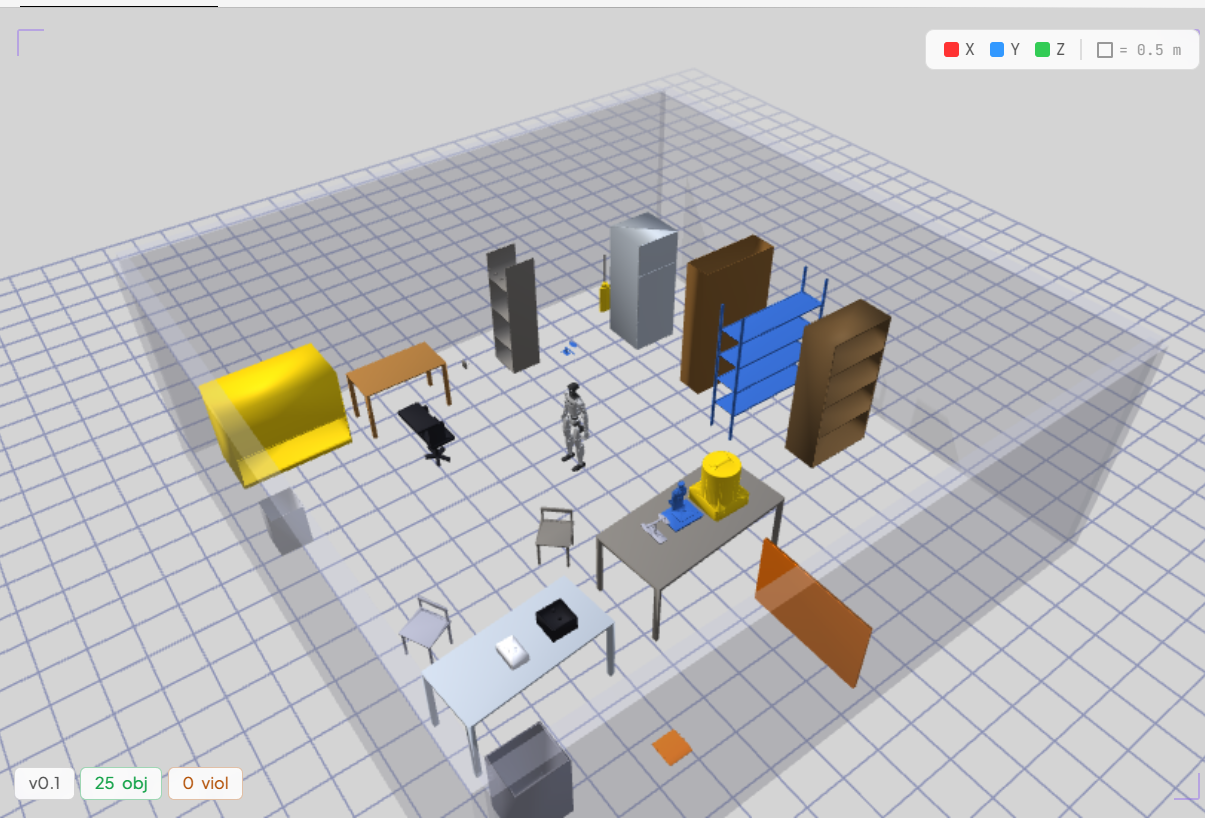}
        \caption{Laboratory}
    \end{subfigure}    
    \caption{Representative generated simulation scenes rendered in the SR-Platform browser viewer. Each scene is produced from a natural-language workspace description and includes generated or retrieved assets, spatial layout, robot placement, and MJCF-compatible geometry for downstream MuJoCo simulation.}
    \label{fig:scene_examples}
\end{figure}

\subsection{Asset Generation Reliability}

Asset-generation reliability was measured from the L2 \texttt{skill\_generator} calls. Among 300 first-attempt asset-generation calls, 34 required at least one retry, corresponding to an 11.3\% first-attempt retry rate. Failures were primarily associated with malformed CadQuery code, invalid geometry, or generated meshes that failed validation. Automatic retry allows these errors to be corrected without aborting the full scene-generation job.

When repeated generation attempts fail, the system falls back to simplified geometry so that MJCF assembly can still complete. This behavior favors pipeline robustness over perfect visual fidelity: a scene can still be generated and inspected even if one asset must be represented approximately.

\subsection{System Throughput}

The deployed worker configuration supports five concurrent scene-generation jobs through \texttt{WORKER\_CONCURRENCY}. Per-user fairness is enforced with \texttt{MAX\_JOBS\_PER\_USER}, preventing a single user from occupying all worker capacity. The queue accepts up to 100 jobs through \texttt{MAX\_QUEUE\_SIZE}.

Throughput is primarily limited by L2 asset generation, since cache-miss objects require LLM calls and CadQuery execution. However, the architecture supports horizontal scaling: additional workers can be launched while sharing the same Redis, PostgreSQL, MinIO, and Qdrant services. As cache hit rates improve, fewer L2 generation calls are required, increasing effective throughput without changing the worker configuration.

\begin{figure}[!ht]
    \centering
        \begin{subfigure}{0.49\linewidth}
        \includegraphics[width=\linewidth]{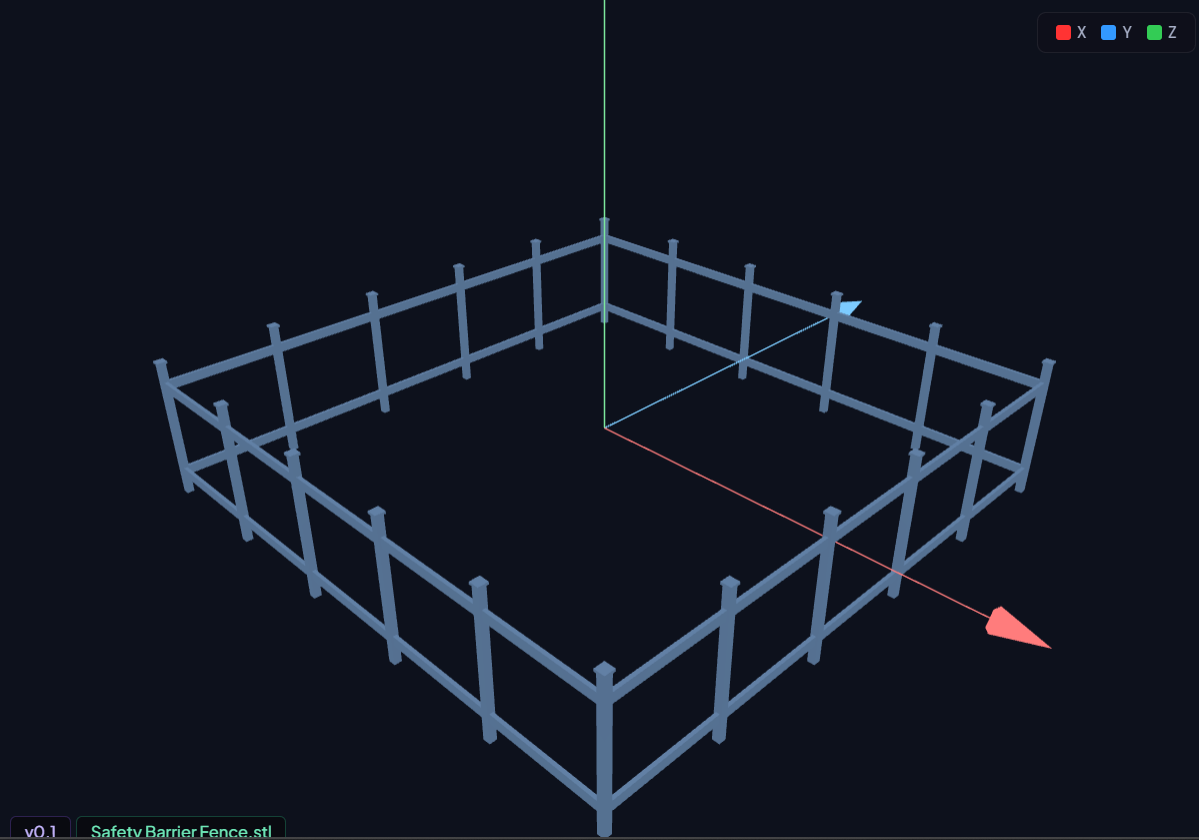}
        \caption{Barrier Fence}
    \end{subfigure}
    \begin{subfigure}{0.49\linewidth}
        \includegraphics[width=\linewidth]{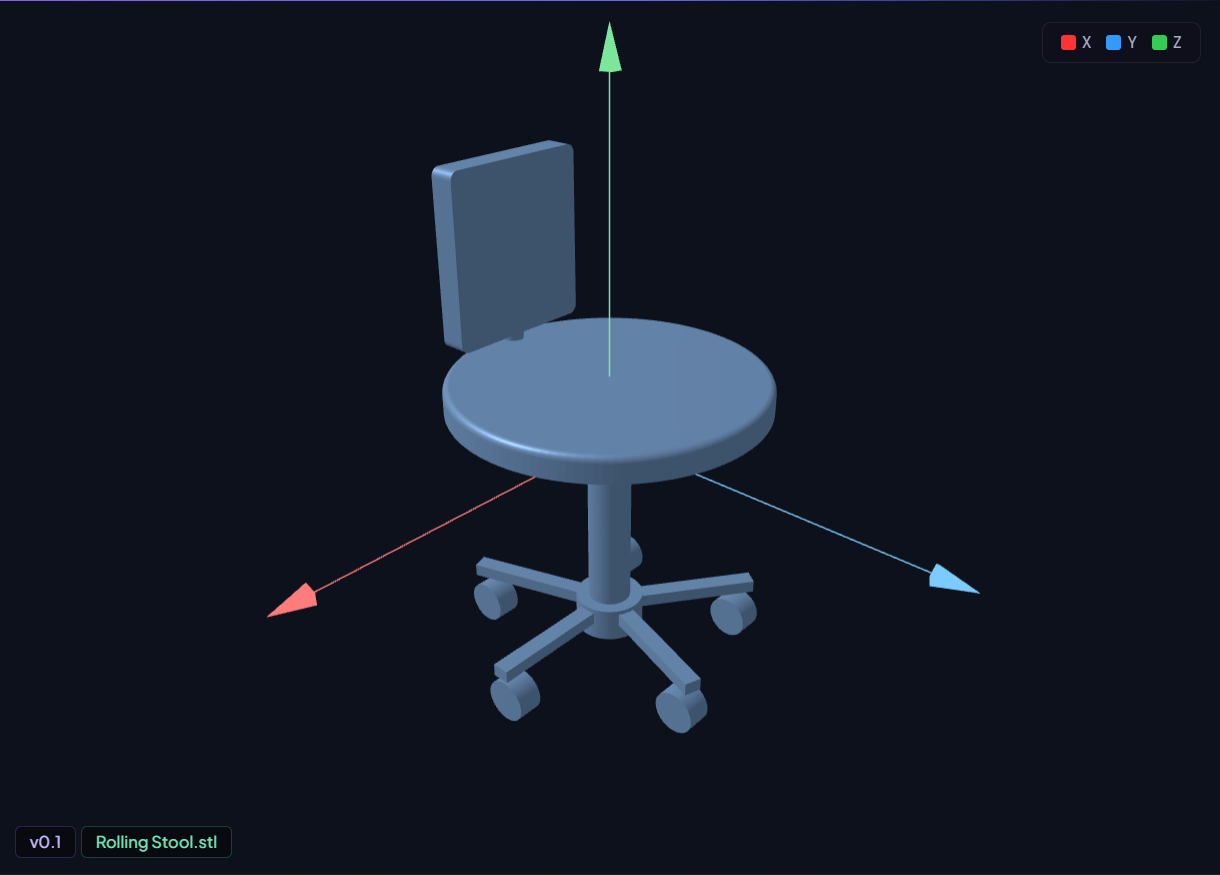}
        \caption{Rolling Stool}
    \end{subfigure}
    \begin{subfigure}{0.49\linewidth}
        \includegraphics[width=\linewidth]{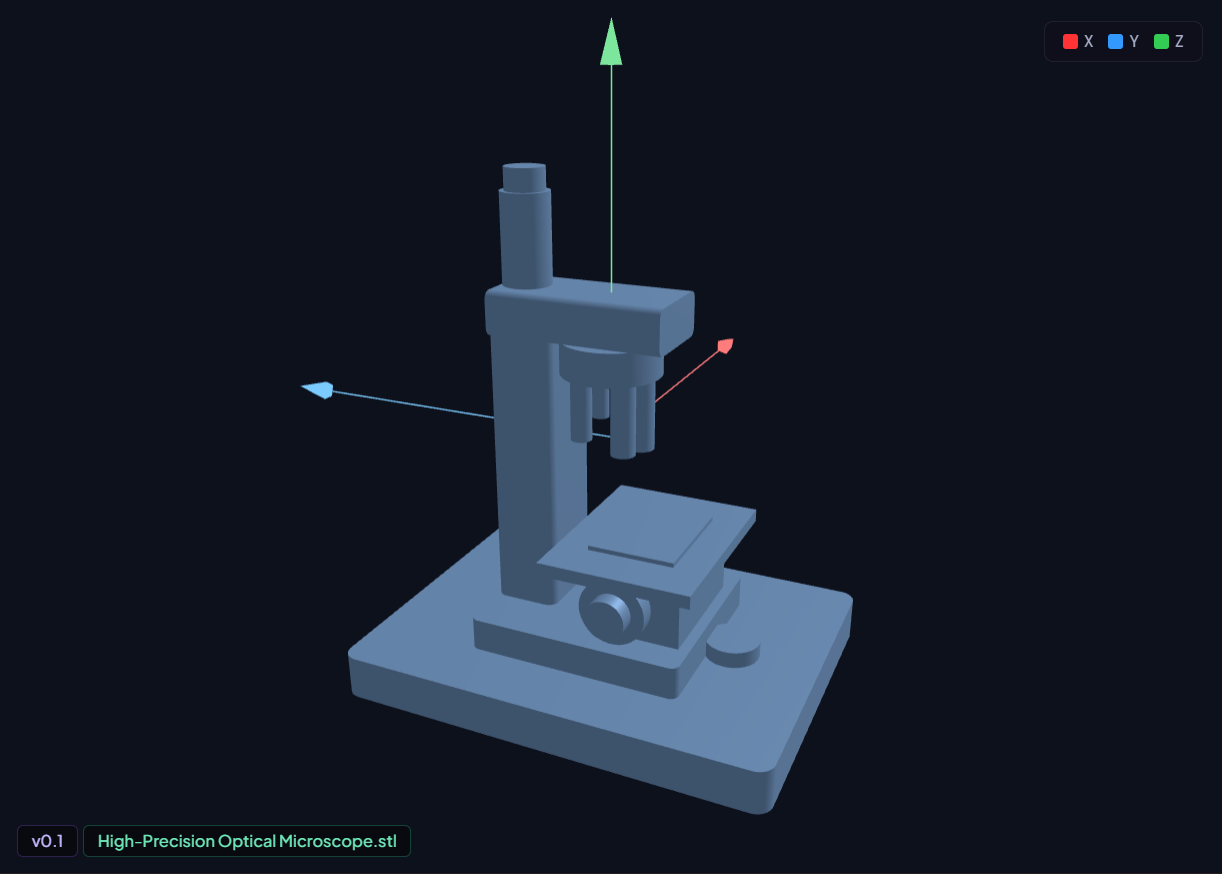}
        \caption{Optical Microscope}
    \end{subfigure}
    \begin{subfigure}{0.49\linewidth}
        \includegraphics[width=\linewidth]{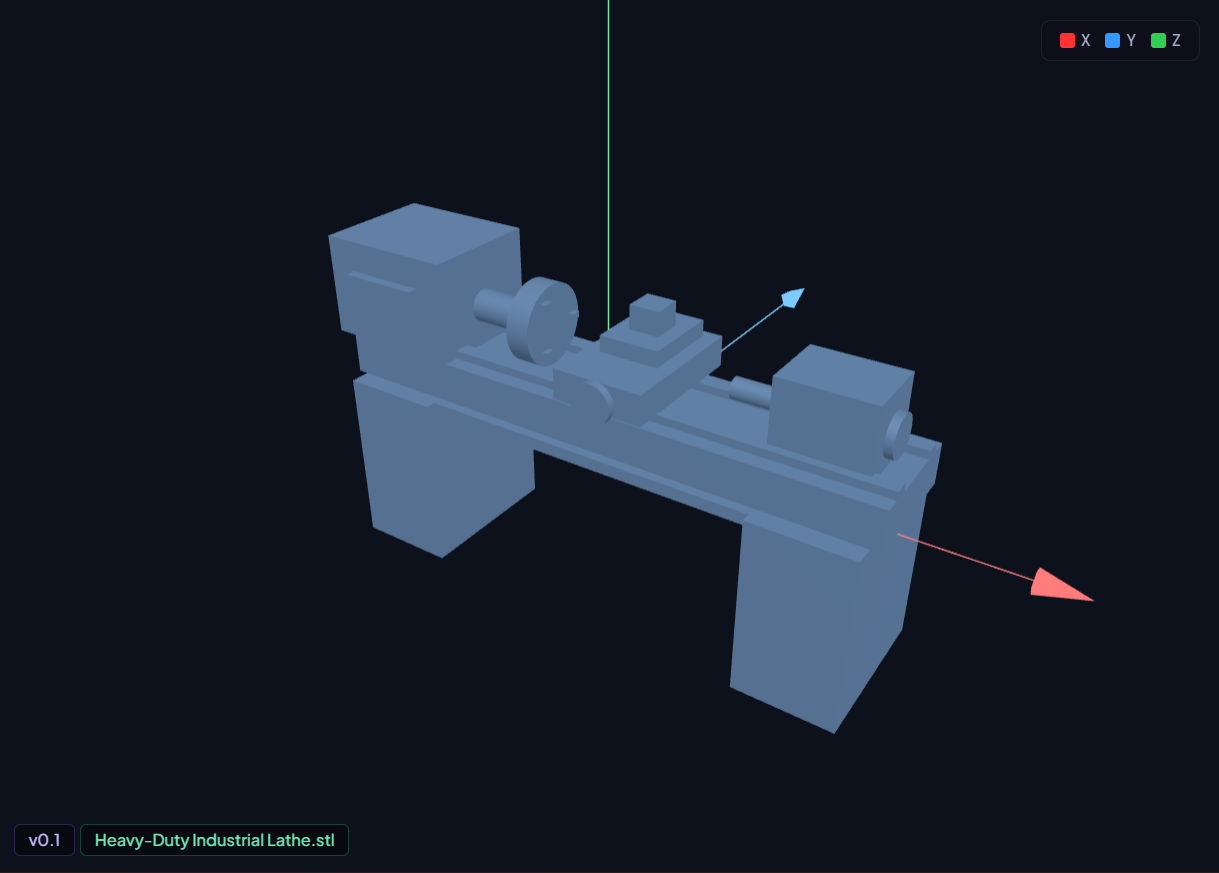}
        \caption{Heavy-Duty Industrial Lathe}
    \end{subfigure}    
    \caption{Asset Studio examples for text-driven 3D asset generation. When semantic retrieval does not find a sufficiently similar cached asset, SR-Platform invokes the asset forge to synthesize CadQuery/OpenSCAD geometry, render it to STL, preview the mesh, and register the result for future reuse.}
    \label{fig:asset_studio}
\end{figure}
\subsection{Qualitative Examples}

Figure~\ref{fig:scene_examples} shows representative simulation scenes generated by SR-Platform from natural-language workspace descriptions. These examples illustrate the complete output of the pipeline: retrieved or synthesized assets, spatial object placement, robot integration, and browser-based MuJoCo visualization.

Figure~\ref{fig:asset_studio} shows examples from the Asset Studio. When semantic retrieval does not find a suitable cached object, the asset-generation pipeline synthesizes CadQuery or OpenSCAD geometry, renders the result as an STL mesh, previews it in the interface, and stores it for future reuse.
%==============================================================================
\section{System Features}

Beyond the core L1--L4 synthesis pipeline, SR-Platform includes several user-facing and production-oriented features that support interactive scene creation, asset reuse, and dataset generation.

\subsection{Qualitative Scene Generation}

Fig.~\ref{fig:scene_examples} shows representative scenes generated by SR-Platform from natural-language workspace descriptions. These examples illustrate the complete output of the pipeline: object assets are retrieved or synthesized, spatial layouts are assigned, robot models are merged into the environment, and the resulting MJCF scene is rendered directly in the browser. The examples demonstrate that the system can generate complete simulation workspaces rather than isolated objects or symbolic scene descriptions.

\subsection{Asset Studio and Interactive Asset Reuse}

SR-Platform provides an Asset Studio for generating, previewing, and reusing 3D objects. Fig.~\ref{fig:asset_studio} shows examples of assets produced through the text-driven geometry generation workflow. When Qdrant retrieval does not find a sufficiently similar cached object, the asset forge synthesizes CadQuery or OpenSCAD geometry, renders the result as an STL mesh, and stores the generated asset for future reuse. Users can also drag assets from the library into an existing scene, after which the backend recomputes the MJCF scene and clamps the object placement within the room boundary.

\subsection{Prompt Refinement and Natural-Language Editing}

A brainstorm chat module allows users to refine scene descriptions through multi-turn interaction before launching generation. This is useful when the initial prompt is underspecified or when the user wants to iteratively add objects, constraints, or task context. After a scene is generated, natural-language object manipulation is also supported. Commands such as ``remove the chair next to the conveyor'' are resolved into object-level scene edits, allowing users to modify the environment without manually editing MJCF XML.

\subsection{Robot Catalog and Scene Merging}

The system includes a robot catalog containing commonly used robot models such as TurtleBot, UR5, Franka Panda, and other supported platforms. A selected robot can be merged into any generated environment through the bridge layer. The robot merger resolves the robot description, inserts the model into the scene, and aligns the coordinate frames so that the final MJCF remains executable in MuJoCo and renderable in the browser.

\subsection{Anomaly Injection and Dataset Export}

SR-Platform supports anomaly injection for generating training data for anomaly detection or robustness evaluation. When enabled, the pipeline can deliberately introduce controlled violations such as object overlap, scale mismatch, or incorrect placement. These anomaly-labeled scenes can be exported together with the MJCF file, mesh assets, and textures as a complete package for offline simulation or downstream robot-learning workflows.

\subsection{Authentication, Persistence, and Observability}

The deployed system includes JWT-based authentication, role-based access control, per-user scene isolation, and persistent job state. Generated scenes are stored per user, while asset metadata and prompt logs are recorded for auditability. Telemetry is written to InfluxDB, enabling monitoring of LLM latency, token usage, retry behavior, and system throughput. These production features are important for operating SR-Platform as a multi-user scene-generation service rather than a single-user research prototype.

%==============================================================================
\section{Discussion}
\label{sec:discussion}

\subsection{Latency and Cache-First Scene Generation}

The production telemetry indicates that SR-Platform is primarily limited by L2 asset generation rather than MJCF assembly or robot merging. As shown in Table~\ref{tab:llm_latency}, cache-miss asset generation requires an LLM-to-CadQuery call for each missing object, making the asset forge the dominant contributor to latency. Table~\ref{tab:e2e_latency} further shows that end-to-end latency is strongly affected by the number of cache-miss objects in a scene.

This result supports the cache-first design of the platform. When an object can be retrieved from Qdrant, the system avoids both the LLM call and the geometry-generation retry loop. As the asset library grows, more object requests can be served through semantic retrieval, reducing latency and improving reliability. In this sense, SR-Platform has a compounding deployment advantage: each successful generation enriches the asset database and can accelerate future generations.

\subsection{Reliability of LLM-Generated Geometry}

The asset forge demonstrates that LLM-generated CadQuery code can be used in a production scene-generation pipeline, but it also exposes clear reliability limits. The observed 11.3\% first-attempt retry rate shows that generated CAD code may be malformed, geometrically invalid, or semantically inconsistent with the requested object. Automatic retry and fallback geometry prevent these failures from aborting the entire scene-generation job, but they do not fully solve the problem of geometric fidelity.

This suggests that LLM-to-CAD generation should be treated as a recoverable subsystem rather than a guaranteed deterministic compiler. In practice, robustness is achieved through validation, retry, caching, and graceful degradation. Future versions should further separate failure categories, including syntax errors, invalid solids, scale errors, topology failures, and semantic mismatches. Such a taxonomy would make it possible to route different failure types to different recovery strategies.

\subsection{Industrial Constraints and Physical Plausibility}

A key design choice in SR-Platform is that layout validity is checked before MJCF assembly. This differs from systems that generate visually plausible scenes without considering deployment constraints. For industrial robot learning, physical plausibility is not limited to collision-free placement; environments should also reflect clearance zones, egress requirements, machinery safety distances, and ventilation constraints.

The current rule checker provides a first step toward this goal by returning structured layout violations. However, these checks should be interpreted as engineering constraints rather than formal certification. NEC, NFPA, ISO, and ASHRAE standards are complex and domain-specific, and a generated scene should not be treated as automatically compliant with all real-world regulations. The practical value of the checker is that it makes constraint violations visible during scene generation, allowing generated environments to better approximate industrial workspaces.

\subsection{Instrumentation Gaps}

Two instrumentation gaps limit the current evaluation. First, L1 orchestration and L3 layout calls are grouped under an \texttt{unknown} tag in the telemetry, preventing a clean per-stage comparison between prompt parsing and layout reasoning. Second, direct wall-clock pipeline-stage timing was not fully connected during the production measurement window. As a result, the end-to-end latency values in Table~\ref{tab:e2e_latency} are estimated from observed LLM-call distributions and known pipeline structure rather than measured as complete request traces.

These gaps do not invalidate the main latency conclusion, because the L2 asset-generation calls are directly measured and clearly dominate runtime. However, future deployments should log stage-level wall-clock timing for L1, L2, L3, and L4 separately. This would enable more precise analysis of queueing delay, asset retrieval time, mesh processing time, robot merge time, and frontend rendering overhead.

\subsection{Limitations}

SR-Platform currently targets indoor industrial and semi-structured manipulation environments. The system is best suited for workspaces composed of objects that can be represented by procedural or mesh-based geometry, such as tables, shelves, conveyors, boxes, panels, fixtures, and machinery components. Highly irregular, organic, deformable, or visually detailed objects remain more difficult for LLM-to-CadQuery synthesis.

The layout architect also has limits. Very large facilities, dense multi-room environments, and scenes requiring fine-grained task-specific affordance reasoning may exceed the current spatial reasoning capability of the pipeline. In addition, the current system generates simulation environments, but it does not yet automatically generate complete robot-control tasks, reward functions, demonstrations, or policy-training datasets.

\subsection{Future Directions}

Several extensions are planned. First, stage-level instrumentation will be added to provide direct wall-clock measurements for every pipeline layer. Second, the asset-generation benchmark will be expanded with a more detailed failure taxonomy and additional model-routing policies. Third, image-to-3D asset generation could be integrated so that users can create simulation assets from photographs of real industrial objects. Fourth, higher-fidelity physics backends could be added for contact-rich manipulation, deformable objects, and cable-like structures. Finally, direct export to robot-learning dataset formats such as LeRobot and RLDS would allow SR-Platform to move beyond scene generation toward full training-data generation.

%==============================================================================
\section{Conclusion}

This paper presented SR-Platform, a production-deployed agentic system for synthesizing executable robot simulation environments from natural-language descriptions. The system converts a free-form workspace prompt into a complete MuJoCo scene containing generated or retrieved assets, spatially assigned object poses, an integrated robot model, and spawn coordinates. By decomposing scene synthesis into orchestration, asset generation, layout reasoning, and MJCF assembly, SR-Platform avoids relying on a single monolithic LLM call and instead uses language models within a structured, auditable pipeline. SR-Platform also demonstrates that production-oriented features are important for practical robot-environment synthesis. Real-time progress streaming, asynchronous job execution, semantic asset retrieval, mesh storage, per-user scene state, robot merging, anomaly injection, and telemetry collection allow the system to operate as more than a research prototype. 

Overall, SR-Platform represents a step toward making robot simulation environment generation more scalable and accessible. By allowing users to create executable MuJoCo scenes from plain English prompts, the system reduces the manual effort required to build diverse training environments and provides a foundation for future synthetic data pipelines for robot learning.

\section*{Appendix}

\subsection*{Benchmark Object List}
\label{app:object-list}

\begin{figure}
    \centering
    \includegraphics[width=0.9\linewidth]{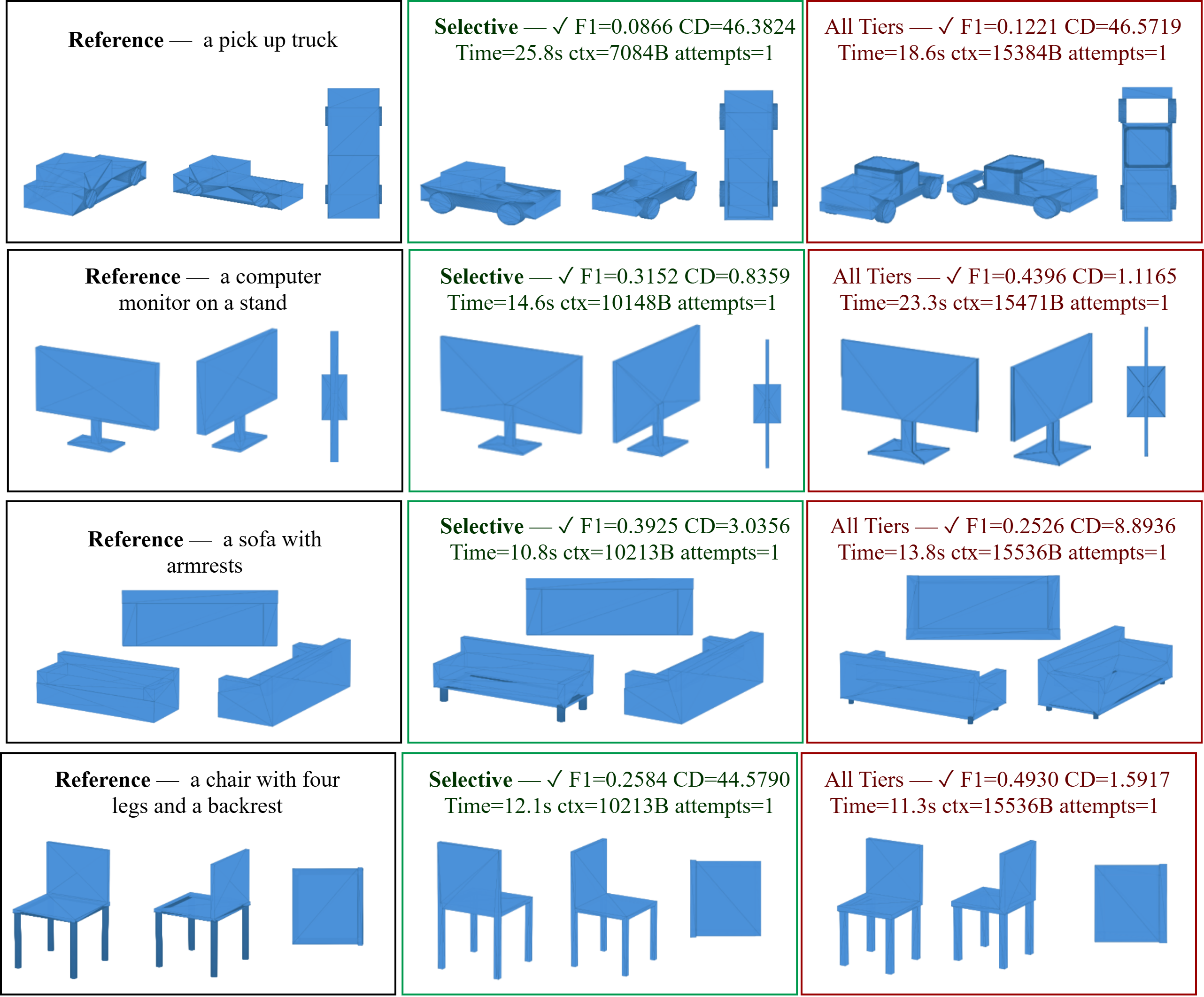}
    \caption{Qualitative asset-generation comparison between reference objects and two prompt-routing configurations. Each row shows the target object, generated outputs, geometric similarity metrics, generation time, context size, and retry count, illustrating how routing strategy affects mesh fidelity and latency across object categories.}
    \label{fig:object_asset}
\end{figure}
The asset-generation benchmark used two object suites: \texttt{abstract\_45}, a 45-object free-form set split evenly across easy, medium, and hard prompts, and \texttt{standard\_100}, a 100-object mechanical-CAD set split into T1 primitive solids, T2 mechanical parts, and T3 compound assemblies. Tables~\ref{tab:abstract45-objects} and~\ref{tab:standard100-objects} list the prompts used for qualitative and quantitative geometry evaluation.

\begingroup
\scriptsize
\setlength{\tabcolsep}{4pt}
\renewcommand{\arraystretch}{1.08}
\begin{longtable}{p{0.22\linewidth}p{0.23\linewidth}p{0.47\linewidth}}
\caption{Benchmark object list for the \texttt{abstract\_45} free-form asset-generation task set. The dataset contains simple, medium, and hard object prompts designed to evaluate visual-semantic asset synthesis beyond precise mechanical CAD primitives.}
\label{tab:abstract45-objects}\\
\toprule
ID & Tier & Object prompt \\
\midrule
\endfirsthead
\toprule
ID & Tier & Object prompt \\
\midrule
\endhead
\midrule
\multicolumn{3}{r}{\footnotesize Continued on next page}\\
\endfoot
\bottomrule
\endlastfoot
\texttt{QA\_001} & easy & a ball \\
\texttt{QA\_002} & easy & a large box \\
\texttt{QA\_003} & easy & a tall bottle shape \\
\texttt{QA\_004} & easy & a thick disc \\
\texttt{QA\_005} & easy & a cone \\
\texttt{QA\_006} & easy & a long hollow pipe \\
\texttt{QA\_007} & easy & a wedge shape \\
\texttt{QA\_008} & easy & a dome \\
\texttt{QA\_009} & easy & a large bowl \\
\texttt{QA\_010} & easy & a thick ring \\
\texttt{QA\_011} & easy & a wide flat board \\
\texttt{QA\_012} & easy & a tall pillar \\
\texttt{QA\_013} & easy & an L-shaped block \\
\texttt{QA\_014} & easy & a plus-shaped block \\
\texttt{QA\_015} & easy & a large bucket shape \\
\texttt{QA\_016} & medium & a coffee mug with a handle \\
\texttt{QA\_017} & medium & a hammer \\
\texttt{QA\_018} & medium & a dumbbell \\
\texttt{QA\_019} & medium & a mushroom shape \\
\texttt{QA\_020} & medium & a saucepan with a handle \\
\texttt{QA\_021} & hard & a trophy cup on a pedestal \\
\texttt{QA\_022} & medium & a flashlight \\
\texttt{QA\_023} & medium & a traffic cone on a flat base \\
\texttt{QA\_024} & medium & a table lamp \\
\texttt{QA\_025} & medium & a goblet or wine glass \\
\texttt{QA\_026} & medium & a toilet plunger \\
\texttt{QA\_027} & hard & a desk fan shape \\
\texttt{QA\_028} & medium & a microphone \\
\texttt{QA\_029} & medium & a simple wrench \\
\texttt{QA\_030} & medium & a mallet \\
\texttt{QA\_031} & hard & a fire extinguisher \\
\texttt{QA\_032} & medium & a chair with four legs and a backrest \\
\texttt{QA\_033} & medium & a dining table with four legs \\
\texttt{QA\_034} & hard & a bookshelf with shelves \\
\texttt{QA\_035} & hard & a desk lamp with a bent arm \\
\texttt{QA\_036} & hard & a staircase with three steps \\
\texttt{QA\_037} & hard & a sofa with armrests \\
\texttt{QA\_038} & hard & a computer monitor on a stand \\
\texttt{QA\_039} & hard & a coat rack with hooks \\
\texttt{QA\_040} & hard & a shopping cart \\
\texttt{QA\_041} & hard & a guitar shape \\
\texttt{QA\_042} & hard & a simple desk with drawers \\
\texttt{QA\_043} & hard & a wheelbarrow \\
\texttt{QA\_044} & hard & a bathtub with legs \\
\texttt{QA\_045} & hard & a pickup truck \\
\end{longtable}
\endgroup

\begingroup
\scriptsize
\setlength{\tabcolsep}{4pt}
\renewcommand{\arraystretch}{1.08}
\begin{longtable}{p{0.12\linewidth}p{0.13\linewidth}p{0.67\linewidth}}
\caption{Benchmark object list for the \texttt{standard\_100} mechanical-CAD task set. The dataset contains dimensioned geometric, structural, pipe, bracket, gear, enclosure, and drone-frame prompts used to evaluate accuracy of LLM-generated CadQuery geometry.}
\label{tab:standard100-objects}\\
\toprule
ID & Tier & Object prompt \\
\midrule
\endfirsthead
\toprule
ID & Tier & Object prompt \\
\midrule
\endhead
\midrule
\multicolumn{3}{r}{\footnotesize Continued on next page}\\
\endfoot
\bottomrule
\endlastfoot
\texttt{T1\_001} & T1 & A rectangular block 50mm long, 30mm wide, and 20mm tall. \\
\texttt{T1\_002} & T1 & A cylinder with a diameter of 25mm and a height of 40mm. \\
\texttt{T1\_003} & T1 & A solid sphere with a radius of 15mm. \\
\texttt{T1\_004} & T1 & A cone with a base diameter of 30mm and a height of 45mm. \\
\texttt{T1\_005} & T1 & A cube with 40mm sides. \\
\texttt{T1\_006} & T1 & A flat rectangular plate, 100mm by 60mm, with a thickness of 3mm. \\
\texttt{T1\_007} & T1 & A hexagonal prism with a circumscribed circle diameter of 20mm and a height of 35mm. \\
\texttt{T1\_008} & T1 & A solid upper hemisphere with a diameter of 50mm, flat face on the XY plane, dome rising in +Z, centered at the origin. \\
\texttt{T1\_009} & T1 & A thin disc, 50mm in diameter and 2mm thick. \\
\texttt{T1\_010} & T1 & A tall cylindrical rod, 8mm in diameter and 150mm long. \\
\texttt{T1\_011} & T1 & A rectangular bar 10mm wide, 10mm tall, and 200mm long. \\
\texttt{T1\_012} & T1 & A triangular prism with an equilateral triangle cross-section having a circumscribed circle diameter of 30mm (centroid at the origin), extruded to 50mm. \\
\texttt{T1\_013} & T1 & A stadium (oblong) shape 60mm long (X), 20mm wide (Y), and 8mm thick (Z), lying flat on the XY plane, centered at the origin. The shape is a rectangle with semicircular ends along the X axis. \\
\texttt{T1\_014} & T1 & A regular octagonal prism with a circumscribed circle diameter of 40mm (i.e., all vertices lie on a 40mm diameter circle), extruded to a height of 15mm. \\
\texttt{T1\_015} & T1 & A truncated cone (frustum) with a base diameter of 40mm, a top diameter of 20mm, and a height of 30mm. \\
\texttt{T1\_016} & T1 & A solid elliptical cylinder with a major axis of 40mm, minor axis of 20mm, and a height of 25mm. \\
\texttt{T1\_017} & T1 & A square plate, 80mm on each side, 5mm thick. \\
\texttt{T1\_018} & T1 & A pentagonal prism with a circumscribed circle diameter of 25mm and a height of 40mm. \\
\texttt{T1\_019} & T1 & A rectangular block with dimensions 120mm by 45mm by 15mm. \\
\texttt{T1\_020} & T1 & A cylinder with a diameter of 36mm and a height of 60mm (Z), centered at the origin on the XY plane. \\
\texttt{T1\_021} & T1 & A solid wedge (triangular prism) centered at the origin. The rectangular base is 60mm (X) by 30mm (Y) on the XY plane. It tapers symmetrically from both Y edges to a thin ridge line at Y=0, running along X for 60mm, at height Z=25mm. \\
\texttt{T1\_022} & T1 & A small cube with 10mm sides. \\
\texttt{T1\_023} & T1 & A rounded rectangular block 50mm long, 25mm wide, and 15mm tall, with 3mm fillets on all vertical edges. \\
\texttt{T1\_024} & T1 & A chamfered cube with 30mm sides and 2mm chamfers on all edges. \\
\texttt{T1\_025} & T1 & A large flat washer: outer diameter 50mm, inner diameter 25mm, thickness 4mm. \\
\texttt{T1\_026} & T1 & A hollow tube with an outer diameter of 30mm, inner diameter of 24mm, and length of 80mm. \\
\texttt{T1\_027} & T1 & A rectangular block 70mm by 40mm by 25mm with 5mm fillets on all edges. \\
\texttt{T1\_028} & T1 & A dome shape centered at the origin on the XY plane. A flat circular base 40mm in diameter and 3mm thick (Z=0 to Z=3), topped by a hemisphere of 40mm diameter (matching the base) centered on the top face of the base. \\
\texttt{T1\_029} & T1 & A solid right-angle triangular prism on the XY plane. The right triangle has a 20mm leg along X and a 30mm leg along Y, with the right angle at the origin. Extruded 40mm in +Z. \\
\texttt{T1\_030} & T1 & A cylinder with a 12mm diameter, 100mm tall, with 1mm chamfers on both circular edges. \\
\texttt{T1\_031} & T1 & A tapered rectangular block (rectangular frustum): the base is 50mm (X) by 30mm (Y) on the XY plane, the top face is 30mm (X) by 15mm (Y) centered above the base, and the height is 40mm along Z. \\
\texttt{T1\_032} & T1 & A regular dodecagonal (12-sided) prism with a circumscribed circle diameter of 35mm and a height of 20mm. \\
\texttt{T1\_033} & T1 & A regular dodecagonal (12-sided) prism with a circumscribed circle diameter of 50mm and a height of 10mm. \\
\texttt{T1\_034} & T1 & A solid cylinder 20mm in diameter and 20mm tall, with 2mm fillets on both top and bottom circular edges. \\
\texttt{T1\_035} & T1 & A box 35mm long, 35mm wide, and 50mm tall. \\
\texttt{T1\_036} & T1 & A flat circular disc with a diameter of 80mm and a thickness of 1mm. \\
\texttt{T1\_037} & T1 & A pyramid with a square base of 40mm sides and a height of 50mm. \\
\texttt{T1\_038} & T1 & A hexagonal bar: hexagonal cross-section with a 15mm circumscribed circle diameter, 120mm long. \\
\texttt{T1\_039} & T1 & A rectangular block 25mm by 15mm by 10mm with 1mm chamfers on all edges. \\
\texttt{T1\_040} & T1 & A coin-shaped disc: 38mm diameter, 3mm thick, with 0.5mm fillets on both circular edges. \\
\texttt{T1\_041} & T1 & A solid ring: outer diameter 40mm, inner diameter 30mm, height 12mm. \\
\texttt{T1\_042} & T1 & A rectangular prism 90mm long, 20mm wide, 8mm tall, with 2mm fillets on all long edges. \\
\texttt{T1\_043} & T1 & A capsule shape: a cylinder 30mm in diameter and 50mm tall, with hemispherical caps on both ends. \\
\texttt{T1\_044} & T1 & A thin rectangular fin: 80mm tall, 60mm long, and 2mm thick. \\
\texttt{T1\_045} & T1 & A semicircular half-pipe trough centered at the origin: the cross-section is a semicircular arc on the XZ plane opening upward (outer radius 25mm, inner radius 22mm, giving 3mm wall thickness). The trough is 50mm wide (X), 25mm tall (Z), and extruded 70mm along the Y axis, centered so it spans Y=-35 to Y=+35. \\
\texttt{T1\_046} & T1 & A cube with 20mm sides and 3mm fillets on all edges. \\
\texttt{T1\_047} & T1 & A truncated cone with a base diameter of 60mm, top diameter of 40mm, and height of 35mm. \\
\texttt{T1\_048} & T1 & A solid rectangular block 150mm long, 50mm wide, and 30mm tall. \\
\texttt{T1\_049} & T1 & A cylinder with a 40mm diameter and 10mm height, resembling a hockey puck. \\
\texttt{T1\_050} & T1 & A solid rectangular block 40mm by 20mm by 60mm, with 4mm fillets on only the four vertical edges. \\
\texttt{T2\_001} & T2 & A flat washer: outer diameter 20mm, inner diameter 10.5mm, thickness 2mm. The washer lies flat on the XY plane, centered at the origin, with thickness extruded in the +Z direction. \\
\texttt{T2\_002} & T2 & A hexagonal nut: a regular hexagonal prism with a 19.6mm circumscribed circle diameter (approximately 17mm across flats), 8mm tall, with a 10mm diameter through-hole centered along the Z axis. The nut sits flat on the XY plane, centered at the origin. \\
\texttt{T2\_003} & T2 & A hex bolt: a 10mm diameter cylindrical shaft, 30mm long, with a hexagonal head on top. The head has a 19.6mm circumscribed circle diameter (approximately 17mm across flats) and is 7mm tall. The bolt stands upright with the shaft along the Z axis starting at Z=0, and the hex head sitting on top of the shaft. Centered at the origin in X and Y. \\
\texttt{T2\_004} & T2 & A socket head cap screw (allen bolt): an 8mm diameter cylindrical shaft, 25mm long, topped with a cylindrical head 13mm in diameter and 8mm tall. A hexagonal socket recess is cut into the top face of the head, 6.93mm across corners (6mm across flats), 4mm deep. The screw stands upright with the shaft along Z starting at Z=0, head on top. Centered at the origin in X and Y. \\
\texttt{T2\_005} & T2 & A structural I-beam: 100mm long along the X axis, 40mm total height along Z, 20mm flange width along Y. Both flanges are 3mm thick and the vertical web is 2mm thick. The I-beam cross-section is in the YZ plane, standing upright, centered at the origin. \\
\texttt{T2\_006} & T2 & An L-bracket: vertical wall 50mm tall (Z), 40mm wide (X), 3mm thick (Y). Horizontal base 40mm deep (Y), 40mm wide (X), 3mm thick (Z). The inside corner of the L is at the origin, with the base extending along +Y on the XY plane and the wall rising along +Z. \\
\texttt{T2\_007} & T2 & A U-channel: 80mm long (X), 40mm wide (Y), 30mm tall (Z), wall thickness 3mm, base thickness 3mm. The channel is open on top and open on both ends along X. The U-shaped cross-section is in the YZ plane, extruded along X, centered at the origin. \\
\texttt{T2\_008} & T2 & A flat mounting plate: 80mm (X) by 60mm (Y) by 5mm thick (Z), centered at the origin. Four M4 clearance holes (4.2mm diameter) drilled through the plate, positioned 8mm inset from each corner. \\
\texttt{T2\_009} & T2 & A flanged bushing standing upright along Z, centered at the origin in XY. The flange is at the bottom: 30mm outer diameter, 3mm thick (Z=0 to Z=3). The cylindrical body extends upward from the flange: 20mm outer diameter, 30mm long (Z=3 to Z=33). A 10mm diameter bore runs through the entire length of the part. \\
\texttt{T2\_010} & T2 & A shaft collar: cylindrical ring with 25mm outer diameter, 15mm bore (inner diameter), and 12mm height, with axis along Z. A 5mm diameter radial set screw hole is drilled through the wall at mid-height (Z=6mm), oriented along the X axis. Centered at the origin. \\
\texttt{T2\_011} & T2 & A dowel pin: solid cylinder 8mm in diameter and 40mm long, standing upright along the Z axis, centered at the origin. Both circular ends have a 1mm 45-degree chamfer. \\
\texttt{T2\_012} & T2 & A plate with counterbored holes: 60mm (X) by 40mm (Y) by 8mm thick (Z), centered at the origin. Four counterbored holes at the corners, each positioned 8mm inset from the nearest edges. Each hole is 5.5mm diameter through the full thickness, with a 10mm diameter counterbore 3mm deep on the top face (+Z). \\
\texttt{T2\_013} & T2 & A round flange plate: a solid disc 60mm in diameter and 8mm thick, lying flat on the XY plane, centered at the origin. A 20mm diameter center bore goes through the full thickness. Six 5.5mm diameter bolt holes on a 45mm pitch circle diameter (22.5mm radius from center), at 0, 60, 120, 180, 240, and 300 degrees from the +X axis, through the full thickness. \\
\texttt{T2\_014} & T2 & A hollow rectangular tube: 80mm long (X), outer cross-section 30mm wide (Y) by 40mm tall (Z), wall thickness 2mm on all four sides. Open on both ends along X. The rectangular cross-section is in the YZ plane, extruded along X, centered at the origin. \\
\texttt{T2\_015} & T2 & A shaft with a keyway: solid cylinder 20mm in diameter, 60mm long, standing upright along the Z axis, centered at the origin. A rectangular keyway slot runs the full 60mm length on the +X side of the shaft. The keyway is 6mm wide (in the Y direction) and 3mm deep from the outer surface, with its flat bottom at X=7mm. \\
\texttt{T2\_016} & T2 & A shoulder screw standing upright along Z, centered at the origin in XY. From bottom to top: a threaded section (simplified as a plain cylinder) 6mm diameter (Z=0 to Z=10), then a smooth shoulder section 10mm diameter (Z=10 to Z=30), then a cylindrical head 14mm diameter (Z=30 to Z=35). Total height 35mm, with the base at Z=0. \\
\texttt{T2\_017} & T2 & A T-beam: 80mm long along X, 40mm total height along Z, flange 30mm wide along Y. The vertical web is 3mm thick (Y) and 37mm tall, rising from Z=0 to Z=37. The horizontal flange is 30mm wide (Y) and 3mm thick, sitting on top of the web from Z=37 to Z=40. The T cross-section is in the YZ plane, extruded along X, centered at the origin. \\
\texttt{T2\_018} & T2 & A simplified spur gear with 16 triangular teeth. The tooth profile is a star polygon with 32 vertices alternating between a tip circle of 40mm diameter (radius 20mm) and a root circle of 32mm diameter (radius 16mm), evenly spaced around the circumference and connected by straight lines. The gear has a 10mm diameter center bore and is 8mm thick. Axis along Z, flat on the XY plane, centered at the origin. \\
\texttt{T2\_019} & T2 & A clevis bracket (fork mount): a solid rectangular block 30mm (X) by 20mm (Y) by 25mm (Z), centered at the origin. A 10mm wide slot (in X) is cut from the top face, 15mm deep, centered on the block and running the full Y width, creating two parallel prongs. A 6mm diameter pin hole is drilled through both prongs in the Y direction, centered at X=0 and at the mid-height of the prongs (Z=5mm). \\
\texttt{T2\_020} & T2 & A flanged L-bracket. The base plate is 60mm wide (X), 40mm deep (Y), and 4mm thick (Z), starting at Z=0. The vertical wall is 40mm wide (X), 30mm tall (Z), and 4mm thick (Y), rising from the Y=0 edge of the base. The base extends 10mm past the wall on each side in X, forming flange ears. Two 6mm diameter holes are drilled through the base plate, one in each flange ear at (X=+25, Y=20) and (X=-25, Y=20). Two 5mm diameter holes are drilled through the vertical wall at (X=+12, Z=19) and (X=-12, Z=19). \\
\texttt{T2\_021} & T2 & A heat sink base plate with parallel cooling fins. The base plate is 60mm long (X), 40mm wide (Y), and 3mm thick (Z), centered at the origin in XY, bottom face at Z=0. Five rectangular fins stand upright on the top face, each 2mm thick (X), 40mm wide (Y, matching the base), and 15mm tall (Z), evenly spaced along X at 12mm intervals (centered at X = -24, -12, 0, 12, 24mm). \\
\texttt{T2\_022} & T2 & A square flat washer: a square plate 24mm x 24mm (X x Y), 3mm thick (Z), with a 10.5mm diameter central through-hole. The washer lies flat on the XY plane, centered at the origin, with thickness extruded in the +Z direction. \\
\texttt{T2\_023} & T2 & A Z-bracket (offset mounting bracket): viewed from the side in the XZ plane, the bracket forms a Z shape. The bottom plate extends from X=0 to X=30, 25mm deep (Y, centered), 3mm thick (Z=0 to Z=3). A vertical web 3mm thick (X, centered at X=0) spans the full height Z=0 to Z=26. The top plate extends from X=-30 to X=0, 25mm deep (Y, centered), 3mm thick (Z=23 to Z=26). Centered in Y at the origin. \\
\texttt{T2\_024} & T2 & A spur gear with a protruding hub, standing upright along Z, centered at the origin. The gear disc uses a star-polygon tooth profile with 16 teeth: tooth tips at 45mm diameter (22.5mm radius), tooth roots at 33.75mm diameter (16.875mm radius), face width 8mm (Z=0 to Z=8). A cylindrical hub 16mm in diameter protrudes from the center, extending from Z=0 to Z=20 (12mm above the gear face). An 8mm diameter bore runs through the full height along Z. \\
\texttt{T2\_025} & T2 & A flanged shaft standing upright along Z, centered at the origin. The flange is a disc 40mm in diameter, 5mm thick, at the base (Z=0 to Z=5). A cylindrical shaft 20mm in diameter and 50mm long rises from the center of the flange (Z=5 to Z=55). Four 6mm diameter bolt holes are drilled through the flange on a 30mm bolt circle diameter (15mm radius from center), positioned at 0, 90, 180, and 270 degrees from the +X axis. \\
\texttt{T3\_001} & T3 & An open-top rectangular container with rounded vertical corners. The outer dimensions are 60mm (X) by 40mm (Y) by 30mm tall (Z). The four vertical edges are filleted with a 5mm radius, giving the container a smooth rounded-rectangle cross- section. The top face (+Z) is open (removed), and the container is shelled inward with 2mm wall thickness, preserving the outer dimensions. The container sits on the XY plane with its base at Z=0, centered at the origin. \\
\texttt{T3\_002} & T3 & A rounded rectangular electronics enclosure body, open on top. Outer dimensions 80mm (X) by 50mm (Y) by 25mm tall (Z). The four vertical edges are rounded with an 8mm fillet radius, giving the enclosure a smooth rounded-rectangle cross- section. The top face (+Z) is removed and the body is shelled inward with 2mm wall thickness, creating a hollow tray with the outer dimensions preserved. Four 3.2mm diameter mounting holes are drilled through the bottom face, positioned in a rectangular pattern 64mm (X) by 34mm (Y) centered at the origin (each hole is 8mm inset from the nearest outer edges). The enclosure sits on the XY plane with its base at Z=0, centered at the origin. \\
\texttt{T3\_003} & T3 & A rectangular electronics enclosure body with four internal screw bosses, open on top. The outer box is 70mm (X) by 50mm (Y) by 30mm tall (Z), centered at the origin, base at Z=0. The four vertical edges are filleted with a 4mm radius. The top face (+Z) is removed and the body is shelled inward with 2mm wall thickness. Four solid cylindrical screw bosses, each 8mm in diameter, stand vertically inside the enclosure from the inner floor (Z=2) to the rim height (Z=30). They are positioned symmetrically at (+/-27, +/-17) mm from the center in XY. A 3mm diameter through-hole runs vertically through each boss and the floor below it. \\
\texttt{T3\_004} & T3 & A V-belt pulley standing upright along the Z axis, centered at the origin in XY. The pulley is 20mm wide (Z=0 to Z=20) with an outer diameter of 60mm and a 12mm diameter central bore. The profile, in the XZ plane (X = radial distance, Z = height), is: straight from the bore (X=6) to the rim (X=30) at Z=0, up to Z=5 at X=30 (bottom flange), diagonally inward to the V-groove root at X=20 and Z=10, diagonally outward to X=30 at Z=15 (top flange), up to Z=20 at X=30, then back to X=6 at Z=20 and closed. This profile is revolved 360 degrees around the Z axis. A keyway slot is cut into the bore on the +X side: 4mm wide (Y direction), 2.5mm deep radially (from the bore surface at X=6 to X=8.5), running the full 20mm width along Z. \\
\texttt{T3\_005} & T3 & A flanged pipe section standing upright along the Z axis, centered at the origin in XY. The overall height is 50mm (Z=0 to Z=50). Build the solid body by stacking three concentric cylinders: a bottom flange disc (50mm outer diameter, Z=0 to Z=5), a pipe wall cylinder (30mm outer diameter, Z=0 to Z=50), and a top flange disc (50mm outer diameter, Z=45 to Z=50), then union them. Cut a 20mm diameter bore through the full length along Z. Finally, drill four 6mm diameter bolt holes through both flanges on a 40mm bolt circle at 0, 90, 180, and 270 degrees, using pushPoints. \\
\texttt{T3\_006} & T3 & A hollow vase (bottle shape), open at the top, standing upright along the Z axis, centered at the origin. The outer profile is defined by straight-line segments in the XZ plane (X = radial distance, Z = height): base at radius 25mm at Z=0, straight wall up to Z=5, outward to a belly radius of 28mm at Z=25, inward to a shoulder radius of 22mm at Z=45, further inward to a neck radius of 10mm at Z=55, then a straight neck up to Z=65. The axis of revolution is at X=0. This profile is revolved 360 degrees around the Z axis to form a solid. The top face (+Z) at the neck opening is then removed and the part is shelled inward with 2mm wall thickness to create the hollow interior. \\
\texttt{T3\_007} & T3 & A rectangular-to-round duct transition adapter, standing upright along the Z axis, centered at the origin in XY. The part has three sections from bottom to top: (1) a rectangular mounting flange 70mm (X) by 50mm (Y), 3mm thick, from Z=0 to Z=3; (2) a smooth lofted transition from a 60mm x 40mm rectangle at Z=3 to a circle of 30mm diameter at Z=53; (3) a cylindrical neck 30mm in diameter extending from Z=53 to Z=63. All sections are centered at the origin in XY. \\
\texttt{T3\_008} & T3 & A spur gear on a stepped shaft, standing upright along the Z axis, centered at the origin in XY. The shaft has three cylindrical sections from bottom to top: a 15mm diameter journal from Z=0 to Z=20, a 20mm diameter gear seat from Z=20 to Z=35, and a 15mm diameter journal from Z=35 to Z=55. A 20-tooth spur gear with a star-polygon tooth profile (40 vertices alternating between tip radius 22mm and root radius 18mm) is integrated at the gear seat section from Z=20 to Z=35 (15mm face width). An 8mm diameter bore runs through the full shaft length along Z. A keyway slot 3mm wide (Y) and 3mm deep (radially, from the bore surface at radius 4mm outward to radius 7mm) is cut through the gear seat section only (Z=20 to Z=35), centered on the +X side. \\
\texttt{T3\_009} & T3 & A compound gear set (two gears of different sizes on a shared hub), standing upright along the Z axis, centered at the origin in XY. The cylindrical hub is 16mm in diameter and 40mm long (Z=0 to Z=40). A large spur gear with 24 teeth (star-polygon profile: 54mm tip diameter (27mm radius), 44mm root diameter (22mm radius)) is located from Z=3 to Z=13 (10mm face width). A smaller spur gear with 14 teeth (star-polygon profile: 32mm tip diameter (16mm radius), 26mm root diameter (13mm radius)) is located from Z=20 to Z=30 (10mm face width). An 8mm diameter bore runs through the full hub length along Z. \\
\texttt{T3\_010} & T3 & A triple gear stack on a single hub, standing upright along the Z axis, centered at the origin in XY. The cylindrical hub is 14mm in diameter and 55mm long (Z=0 to Z=55). Three spur gears of decreasing size are spaced along the hub, each with a star-polygon tooth profile: (1) a 24-tooth gear (54mm tip diameter (27mm radius), 44mm root diameter (22mm radius)) from Z=2 to Z=10 (8mm face width); (2) an 18-tooth gear (42mm tip diameter (21mm radius), 34mm root diameter (17mm radius)) from Z=18 to Z=26; (3) a 12-tooth gear (30mm tip diameter (15mm radius), 24mm root diameter (12mm radius)) from Z=34 to Z=42. A 6mm diameter bore runs through the full hub length along Z. A keyway slot 3mm wide (Y) and 2mm deep (radially, from bore surface at radius 3mm to radius 5mm) runs the full 55mm hub length on the +X side. \\
\texttt{T3\_011} & T3 & A spur gear with an extended hub and radial set screw hole, standing upright along the Z axis, centered at the origin in XY. The gear has 20 teeth with a star-polygon tooth profile (48mm tip diameter (24mm radius), 40mm root diameter (20mm radius)) and is 10mm thick (Z=0 to Z=10). A cylindrical hub 16mm in diameter extends from Z=0 to Z=30, protruding 20mm above the gear face. A 10mm diameter bore runs through the full 30mm length along Z. A keyway slot 4mm wide (Y) and 2.5mm deep (radially, from the bore surface at radius 5mm to radius 7.5mm) runs the full 30mm length on the +X side. A 4mm diameter set screw hole is drilled radially through the hub at Z=20mm, oriented along the X axis, passing through both walls of the hub. \\
\texttt{T3\_012} & T3 & An idler gear assembly: a spur gear mounted on a stepped post rising from a square base plate, standing upright along the Z axis, centered at the origin in XY. The base plate is 60mm x 60mm (X x Y), 5mm thick (Z=0 to Z=5). A cylindrical post 20mm in diameter rises from the center of the plate from Z=5 to Z=18. A narrower retaining shoulder 14mm in diameter continues from Z=18 to Z=40. A 16-tooth spur gear with a star-polygon tooth profile (40mm tip diameter (20mm radius), 32mm root diameter (16mm radius)) sits on the post from Z=10 to Z=18 (8mm face width). An 8mm diameter bore runs vertically through the entire assembly along Z. Four 6mm diameter mounting holes are drilled through the base plate at positions (+/-22, +/-22) mm from the center. \\
\texttt{T3\_013} & T3 & A 90-degree pipe elbow with the bend in the XZ plane. The pipe has a 20mm outer diameter and 16mm inner diameter (2mm wall thickness). The bend centerline follows a quarter-circle arc of 40mm radius on the XZ plane: starting at the origin heading in the +Z direction, curving through the first quadrant, and ending at (X=40, Z=40) heading in the +X direction. The pipe cross-section is centered on this arc. The part is centered at Y=0. Created by sweeping the annular pipe cross- section along the arc path defined by a radiusArc from (0,0) to (40,40) with radius 40 on the XZ workplane. \\
\texttt{T3\_014} & T3 & A 180-degree U-bend (return bend) pipe fitting with the bend in the XZ plane. The pipe has a 20mm outer diameter and 16mm inner diameter (2mm wall thickness). The bend centerline follows a semicircular arc of 35mm radius on the XZ plane: starting at the origin heading in the +Z direction, arcing upward through Z=35 at the apex, and returning down to (X=70, Z=0) heading in the -Z direction. The two pipe legs are parallel, 70mm apart in the X direction. The part is centered at Y=0. Created by sweeping the pipe cross-section along the arc path defined by a radiusArc from (0,0) to (70,0) with radius 35 on the XZ workplane. \\
\texttt{T3\_015} & T3 & A pipe tee fitting. The main (run) pipe is a hollow cylinder 30mm outer diameter, 24mm inner diameter (3mm wall thickness), 100mm long, centered at the origin along the X axis (from X=-50 to X=50). A branch pipe of the same diameter exits perpendicular to the main pipe, going in the +Z direction from the intersection at Z=0 up to Z=50. The branch intersects the main pipe at the center, creating a continuous hollow interior shaped like a T. The part is centered at the origin in X and Y. \\
\texttt{T3\_016} & T3 & A 45-degree pipe elbow with the bend in the XZ plane. The pipe has a 20mm outer diameter and 16mm inner diameter (2mm wall thickness). The bend centerline follows a 45-degree arc of 40mm radius on the XZ plane: starting at the origin heading in the +Z direction and curving 45 degrees toward the +X direction. The arc endpoint is at approximately (X=11.72, Z=28.28). The part is centered at Y=0. Created by sweeping the pipe cross-section along the arc path defined by a radiusArc on the XZ workplane with a 40mm bend radius. \\
\texttt{T3\_017} & T3 & A pipe cross fitting (four-way intersection). Two hollow pipes of equal size cross perpendicularly. The first pipe is 30mm outer diameter, 24mm inner diameter (3mm wall thickness), 100mm long along the X axis, centered at the origin (from X=-50 to X=50). The second pipe has the same dimensions, 100mm long along the Z axis, centered at the origin (from Z=-50 to Z=50). The two pipes intersect at the origin, creating a continuous hollow + shaped interior. The part is centered at the origin in all axes. \\
\texttt{T3\_018} & T3 & A quadcopter drone chassis in a plus (+) configuration, lying flat on the XY plane (Z=0 to Z=3), centered at the origin. The central hub is a 30mm x 30mm square plate, 3mm thick. Four rectangular arms extend outward from the hub edges along the +/-X and +/-Y axes: each arm is 45mm long and 10mm wide, running from the hub edge (15mm from center) to 60mm from center. A circular motor mount platform 18mm in diameter is at each arm tip, centered at (+/-60, 0) and (0, +/-60) mm. A 5mm diameter motor shaft hole is drilled through each motor mount. Four M3 flight-controller mounting holes (3.2mm diameter) are drilled through the hub in a 20mm x 20mm square pattern centered at the origin. \\
\texttt{T3\_019} & T3 & A quadcopter drone chassis in an X configuration, lying flat on the XY plane (Z=0 to Z=3), centered at the origin. The central hub is a solid disc 36mm in diameter, 3mm thick. Four rectangular arms extend outward from the hub at 45, 135, 225, and 315 degrees from the +X axis. Each arm is 55mm long and 10mm wide, starting from the hub edge (18mm from center) and reaching 73mm from center. The arms are aligned along their respective diagonal directions using rotated workplanes. A circular motor mount platform 18mm in diameter sits at each arm tip, centered 73mm from the origin along each diagonal (at approximately (+/-51.6, +/-51.6) mm). A 5mm diameter motor shaft hole is drilled through each mount. Four M3 mounting holes (3.2mm diameter) are drilled through the hub in a 20mm x 20mm square pattern centered at the origin. \\
\texttt{T3\_020} & T3 & A quadcopter drone chassis in an H-frame configuration, lying flat on the XY plane (Z=0 to Z=3), centered at the origin. The frame has two parallel side rails running along X, a center plate, four circular motor mounts, and eight drilled holes. Side rails: each rail is 160mm long (X), 12mm wide (Y), 3mm thick (Z). The left rail is centered at Y=-30mm, the right rail at Y=+30mm. Center plate: 50mm long (X), 72mm wide (Y), 3mm thick, centered at the origin, bridging the two rails. Motor mounts: four circular platforms 22mm in diameter, 3mm thick, at the ends of each rail at positions (+80,+30), (+80,-30), (-80,+30), (-80,-30) mm. Holes: one 5mm diameter motor shaft hole drilled through each of the four motor mounts (4 holes total). Four M3 mounting holes (3.2mm diameter) drilled through the center plate at positions (+10,+10), (+10,-10), (-10,+10), (-10,-10) mm. \\
\texttt{T3\_021} & T3 & A pipe flange disc lying flat on the XY plane (Z=0 to Z=16), centered at the origin. The flange is a solid disc 80mm in diameter and 16mm thick. A 32mm diameter center bore passes through the full thickness. Six 9mm diameter bolt holes on a 55mm pitch circle diameter (27.5mm radius from center), at 0, 60, 120, 180, 240, and 300 degrees from the +X axis, drilled through the full thickness. On the top face (Z=16), a concentric gasket groove is machined 2mm deep: the groove is a ring between 38mm inner diameter and 50mm outer diameter, cut from Z=14 to Z=16. \\
\texttt{T3\_022} & T3 & A spoked handwheel lying flat on the XY plane, centered at the origin. The hub is a solid cylinder 30mm in diameter and 12mm tall (Z=0 to Z=12). The outer rim is a ring 100mm outer diameter and 88mm inner diameter (6mm radial width), 8mm tall (Z=0 to Z=8). Four rectangular spokes connect the hub to the rim at 0, 90, 180, and 270 degrees from the +X axis. Each spoke is 8mm wide, 8mm tall (same height as rim), and long enough to bridge from the hub outer edge (15mm from center) to the rim inner edge (44mm from center), centered radially at 29.5mm from the origin. The spokes at 0 and 180 degrees are aligned along X; those at 90 and 270 degrees are aligned along Y. A 14mm diameter center bore passes through the full hub height. \\
\texttt{T3\_023} & T3 & A rectangular mounting plate with six standoff posts, lying on the XY plane (Z=0 to Z=15), centered at the origin. The base plate is 80mm long (X), 50mm wide (Y), and 3mm thick (Z=0 to Z=3). On the top face of the plate, a 3x2 rectangular array of cylindrical standoff posts is arranged with 30mm spacing in X and 30mm spacing in Y, centered on the plate. Each post is 8mm in diameter and 12mm tall (Z=3 to Z=15). A 3.2mm diameter pilot hole (M3 clearance) is drilled through the center of each post and through the plate below it, passing through the full 15mm height. \\
\texttt{T3\_024} & T3 & A flywheel with lightening holes and a raised hub boss, centered at the origin on the XY plane. The main disc is 120mm in diameter and 20mm thick (Z=0 to Z=20). A raised cylindrical hub boss 40mm in diameter and 10mm tall sits on top of the disc, centered at the origin (Z=20 to Z=30). Six circular lightening holes, each 22mm in diameter, are equally spaced on a 70mm pitch circle diameter (35mm radius from center) and pass through the full 20mm disc thickness. A 16mm diameter center bore passes through the entire part height (Z=0 to Z=30). The lightening hole centers are at 0, 60, 120, 180, 240, and 300 degrees from the +X axis. \\
\texttt{T3\_025} & T3 & A flanged shaft coupling standing along the Z axis, centered at the origin in X and Y. The coupling consists of three coaxial cylindrical sections: a bottom flange disc 50mm in diameter and 10mm thick (Z=0 to Z=10), a central hub cylinder 28mm in diameter and 40mm long (Z=10 to Z=50), and a top flange disc 50mm in diameter and 10mm thick (Z=50 to Z=60). The total length is 60mm. Six bolt holes 6.5mm in diameter are equally spaced on a 38mm pitch circle diameter (19mm radius), computed at 0, 60, 120, 180, 240, and 300 degrees from the +X axis. The bolt holes pass through the full 60mm length but only cut the flanges since the hub is narrower. A 14mm diameter center bore runs the full 60mm length. A rectangular keyway slot 5mm wide (Y) and 2.5mm deep (radial, in X) is cut along the bore wall on the +X side, running the full 60mm length. The keyway is centered at X = (bore radius - half keyway depth) = 5.75mm from the origin. \\
\end{longtable}
\endgroup

\subsection*{AI Model Performance for Asset Generation}
\label{app:model-performance}

We benchmarked LLM backends for the CadQuery asset-generation stage on two task families. The composite score is reported on a 0--100 scale and weights latency, success rate, and geometric accuracy. The routing report assigns 40\% of the score to speed and 15\% to success rate; geometric accuracy is measured by Chamfer Distance (CD), using a robust aggregate of $0.6\times$ median CD + $0.4\times$ mean CD. Lower CD indicates closer geometric agreement with the reference object. Costs are normalized as USD per one million input or output tokens.

\begingroup
\scriptsize
\setlength{\tabcolsep}{3pt}
\renewcommand{\arraystretch}{1.08}
\begin{longtable}{p{0.38\linewidth}rrrrr}
\caption{Top-performing LLM backends on the \texttt{standard\_100} mechanical-CAD benchmark. The table reports composite score, average latency, Chamfer Distance median/mean, and input/output token cost, supporting model routing decisions for precision asset generation.}
\label{tab:model-standard100}\\
\toprule
Model & Score & Lat. (s) & CD Med/Mean & In \$/1M & Out \$/1M \\
\midrule
\endfirsthead
\toprule
Model & Score & Lat. (s) & CD Med/Mean & In \$/1M & Out \$/1M \\
\midrule
\endhead
\bottomrule
\endlastfoot
\texttt{claude-opus-4-6-fast} & 86.8 & 8.3 & 0.15 / 1.12 & 36.00 & 180.00 \\
\texttt{qwen3-coder-480b-a35b-instruct-turbo} & 83.1 & 4.3 & 0.16 / 2.45 & 0.35 & 1.50 \\
\texttt{claude-sonnet-4-6} & 79.7 & 22.6 & 0.15 / 0.52 & 3.60 & 18.00 \\
\texttt{claude-opus-4-6} & 79.5 & 17.0 & 0.15 / 1.31 & 6.00 & 30.00 \\
\texttt{openai-gpt-53-codex} & 79.4 & 15.6 & 0.15 / 1.26 & 2.19 & 17.50 \\
\texttt{claude-opus-4-7} & 79.3 & 7.2 & 0.15 / 3.38 & 6.00 & 30.00 \\
\texttt{gemini-3-flash-preview (Native API)} & 78.8 & 12.7 & 0.15 / 1.89 & 0.50 & 3.00 \\
\texttt{gemini-3-1-pro-preview} & 77.4 & 13.3 & 0.15 / 2.09 & 2.50 & 15.00 \\
\texttt{openai-gpt-52} & 76.8 & 19.5 & 0.15 / 1.24 & 2.19 & 17.50 \\
\texttt{deepseek-v4-flash} & 76.5 & 17.2 & 0.16 / 0.98 & 0.17 & 0.35 \\
\texttt{z-ai-glm-5-turbo} & 76.1 & 22.7 & 0.15 / 0.60 & 1.20 & 4.00 \\
\texttt{openai-gpt-4o-2024-11-20} & 75.1 & 4.8 & 0.16 / 5.43 & 3.13 & 12.50 \\
\end{longtable}
\endgroup

\begingroup
\scriptsize
\setlength{\tabcolsep}{3pt}
\renewcommand{\arraystretch}{1.08}
\begin{longtable}{p{0.38\linewidth}rrrrr}
\caption{Top-performing LLM backends on the \texttt{abstract\_45} free-form benchmark. The table reports composite score, average latency, Chamfer Distance median/mean, and input/output token cost, supporting model routing decisions for visually grounded conceptual asset generation.}
\label{tab:model-abstract45}\\
\toprule
Model & Score & Lat. (s) & CD Med/Mean & In \$/1M & Out \$/1M \\
\midrule
\endfirsthead
\toprule
Model & Score & Lat. (s) & CD Med/Mean & In \$/1M & Out \$/1M \\
\midrule
\endhead
\bottomrule
\endlastfoot
\texttt{claude-opus-4-7} & 85.1 & 6.7 & 6.32 / 22.67 & 6.00 & 30.00 \\
\texttt{qwen3-coder-480b-a35b-instruct-turbo} & 81.5 & 8.2 & 13.66 / 27.33 & 0.35 & 1.50 \\
\texttt{gemini-3-1-pro-preview} & 81.5 & 21.2 & 2.52 / 13.59 & 2.50 & 15.00 \\
\texttt{openai-gpt-4o-2024-11-20} & 80.6 & 7.4 & 15.85 / 40.71 & 3.13 & 12.50 \\
\texttt{claude-opus-4-6-fast} & 78.2 & 28.5 & 6.63 / 19.30 & 36.00 & 180.00 \\
\texttt{openai-gpt-53-codex} & 76.1 & 27.9 & 4.57 / 21.25 & 2.19 & 17.50 \\
\texttt{deepseek-v4-flash} & 73.8 & 23.2 & 5.89 / 22.22 & 0.17 & 0.35 \\
\texttt{openai-gpt-4o-mini-2024-07-18} & 72.7 & 5.4 & 17.53 / 35.34 & 0.19 & 0.75 \\
\texttt{qwen3-5-35b-a3b} & 72.2 & 12.4 & 13.05 / 33.86 & 0.31 & 1.25 \\
\texttt{claude-opus-4-5} & 71.2 & 30.4 & 5.31 / 21.94 & 6.00 & 30.00 \\
\texttt{minimax-m27} & 70.5 & 29.3 & 9.51 / 31.66 & 0.38 & 1.50 \\
\texttt{openai-gpt-52-codex} & 69.4 & 36.8 & 6.30 / 20.89 & 2.19 & 17.50 \\
\end{longtable}
\endgroup

\paragraph{Routing recommendation.}
For precision mechanical CAD, \texttt{claude-opus-4-6-fast} achieved the highest score (86.8) but at substantially higher token cost. \texttt{qwen3-coder-480b} was the strongest cost-performance candidate, scoring 83.1 on \texttt{standard\_100} with 4.3\,s average latency and low token cost. For free-form object prompts, \texttt{claude-opus-4-7} ranked first (85.1), while \texttt{qwen3-coder-480b} remained the strongest open-weight routing option.

\paragraph{Native API versus Venice proxy.}
For \texttt{gemini-3-flash-preview} on \texttt{standard\_100}, the Native API scored 78.8 with 12.7\,s average latency and CD 0.15/1.89, while the Venice proxy scored 71.7 with 18.4\,s average latency and CD 0.16/2.58. This indicates that geometric quality was mostly preserved through the proxy path, but routing through the proxy added approximately 6\,s of latency in this benchmark.

{\small
\bibliographystyle{plainnat}
\bibliography{SRv1}

@inproceedings{todorov2012mujoco,
  title={{MuJoCo}: A physics engine for model-based control},
  author={Todorov, Emanuel and Erez, Tom and Tassa, Yuval},
  booktitle={IEEE/RSJ International Conference on Intelligent Robots and Systems (IROS)},
  pages={5026--5033},
  year={2012},
  organization={IEEE}
}

@article{chen2021evaluating,
  title={Evaluating large language models trained on code},
  author={Chen, Mark and Tworek, Jerry and Jun, Heewoo and Yuan, Qiming and Pinto, Henrique Ponde De Oliveira and Kaplan, Jared and Edwards, Harri and Burda, Yuri and Joseph, Nicholas and Brockman, Greg and others},
  journal={arXiv preprint arXiv:2107.03374},
  year={2021}
}

@article{brown2020language,
  title={Language models are few-shot learners},
  author={Brown, Tom and Mann, Benjamin and Ryder, Nick and others},
  journal={Advances in Neural Information Processing Systems},
  volume={33},
  pages={1877--1901},
  year={2020}
}

@inproceedings{ahn2022saycan,
  title={Do as i can, not as i say: Grounding language in robotic affordances},
  author={Ahn, Michael and Brohan, Anthony and Brown, Noah and others},
  booktitle={Conference on Robot Learning (CoRL)},
  year={2022}
}

@article{langchain2024langgraph,
  title={LangGraph vs LangChain: Which Framework Is Best for AI Agents?},
  author={Lei, Zhang},
  journal={MetaVision Journal of Multidisciplinary Studies (MVJMS)},
  volume={1},
  number={4},
  pages={16--26},
  year={2024}
}

@article{yao2023react,
  title={React: Synergizing reasoning and acting in language models},
  author={Yao, Shunyu and Zhao, Jeffrey and Yu, Dian and Du, Nan and Shafran, Izhak and Narasimhan, Karthik and Cao, Yuan},
  journal={arXiv preprint arXiv:2210.03629},
  year={2022}
}

@article{driess2023palm,
  title={Palm-e: An embodied multimodal language model},
  author={Driess, Danny and Xia, Fei and Sajjadi, Mehdi SM and Lynch, Corey and Chowdhery, Aakanksha and Ichter, Brian and Wahid, Ayzaan and Tompson, Jonathan and Vuong, Quan and Yu, Tianhe and others},
  journal={arXiv preprint arXiv:2303.03378},
  year={2023}
}

@inproceedings{matas2018sim,
  title={Sim-to-real reinforcement learning for deformable object manipulation},
  author={Matas, Jan and James, Stephen and Davison, Andrew J},
  booktitle={Conference on Robot Learning (CoRL)},
  year={2018}
}

@misc{nvidia2026newton,
author = {{The Newton Contributors}},
license = {Apache-2.0},
month = apr,
title = {{Newton: GPU-accelerated physics simulation for robotics and simulation research}},
url = {https://github.com/newton-physics/newton},
year = {2025}
}

@inproceedings{zhao2020sim,
  title={Sim-to-real transfer in deep reinforcement learning for robotics: a survey},
  author={Zhao, Wenshuai and Queralta, Jorge Pe{\~n}a and Westerlund, Tomi},
  booktitle={IEEE Symposium Series on Computational Intelligence (SSCI)},
  year={2020}
}

@inproceedings{tobin2017domain,
  title={Domain randomization for transferring deep neural networks from simulation to the real world},
  author={Tobin, Josh and Fong, Rachel and Ray, Alex and others},
  booktitle={IEEE/RSJ International Conference on Intelligent Robots and Systems (IROS)},
  year={2017}
}

@article{wang2023robogen,
  title={{RoboGen}: Towards unleashing infinite data for automated robot learning via generative simulation},
  author={Wang, Yufei and Jiang, Zhou and Chen, Feng and others},
  journal={arXiv preprint arXiv:2311.01455},
  year={2023}
}

@inproceedings{gan2020threedworld,
  title={{ThreeDWorld}: A platform for interactive multi-modal physical simulation},
  author={Gan, Chuang and Schwartz, Jeremy and Alter, Seth and others},
  booktitle={Advances in Neural Information Processing Systems},
  year={2020}
}

@inproceedings{li2023behavior,
  title={{BEHAVIOR-1K}: A benchmark for embodied AI with 1000 everyday activities and realistic simulation},
  author={Li, Chengshu and Zhang, Ruohan and Wong, Josiah and others},
  booktitle={Conference on Robot Learning (CoRL)},
  year={2023}
}

@misc{cadquery2023,
  title={{CadQuery}: An intuitive, easy-to-use Python module for building parametric 3D CAD models},
  author={{CadQuery Contributors}},
  year={2023},
  howpublished={\url{https://cadquery.readthedocs.io}}
}

@article{poole2022dreamfusion,
  title={{DreamFusion}: Text-to-3D using 2D diffusion},
  author={Poole, Ben and Jain, Ajay and Barron, Jonathan T and Mildenhall, Ben},
  journal={arXiv preprint arXiv:2209.14988},
  year={2022}
}

@article{ma2023eureka,
  title={{Eureka}: Human-level reward design via coding large language models},
  author={Ma, Yecheng Jason and Liang, William and Wang, Guanzhi and others},
  journal={arXiv preprint arXiv:2310.12931},
  year={2023}
}

@inproceedings{liang2023code,
  title={Code as policies: Language model programs for embodied control},
  author={Liang, Jacky and Huang, Wenlong and Xia, Fei and Xu, Peng and Hausman, Karol and Ichter, Brian and Florence, Pete and Zeng, Andy},
  booktitle={2023 IEEE International conference on robotics and automation (ICRA)},
  pages={9493--9500},
  year={2023},
  organization={IEEE}
}

@article{wang2023voyager,
  title={{Voyager}: An open-ended embodied agent with large language models},
  author={Wang, Guanzhi and Xie, Yuqi and Jiang, Yunfan and others},
  journal={arXiv preprint arXiv:2305.16291},
  year={2023}
}

@article{mittal2023orbit,
  title={{Orbit}: A unified simulation framework for interactive robot learning environments},
  author={Mittal, Mayank and Yu, Calvin and Yu, Qinxi and others},
  journal={IEEE Robotics and Automation Letters},
  year={2023}
}
}

\end{document}